\documentclass[manuscript,screen]{acmart}

\AtBeginDocument{%
  \providecommand\BibTeX{{%
    \normalfont B\kern-0.5em{\scshape i\kern-0.25em b}\kern-0.8em\TeX}}}

\setcopyright{none}
\copyrightyear{2023}
\acmYear{2023}
\acmDOI{10.1145/xxxxx.xxxxxx}

\acmBooktitle{The 6th ACM Conference on Fairness, Accountability, and Transparency} 

\usepackage{comment}
\usepackage{amsmath,amsfonts}
\usepackage{graphicx}
\usepackage{textcomp}
\usepackage{xcolor}
\usepackage{subfigure}
\usepackage{balance}
\usepackage{diagbox}
\usepackage{threeparttable}
\usepackage{multirow}
\usepackage{color}
\usepackage{algpseudocode}
\usepackage[ruled]{algorithm2e}

\usepackage{colortbl}
\usepackage{graphicx}
\usepackage{makecell}
\usepackage{wrapfig}

\usepackage{url}
\usepackage{bbold}

\usepackage{setspace}

\begin{document}


\title[\resizebox{3.5in}{!}{Preventing Discriminatory Decision-making in Evolving Data Streams}]{Preventing Discriminatory Decision-making in Evolving Data Streams}


%


\author{Zichong Wang}
\email{zichongw@mtu.edu}
\affiliation{%
  \institution{Michigan Technological University}
  \city{Houghton}
  \state{Michigan}
  \country{USA}
  \postcode{49931}
}

\author{Nripsuta Saxena}
\email{nsaxena@usc.edu}
\affiliation{%
  \institution{University of Southern California}
  \city{Los Angeles}
  \state{California}
  \country{USA}
  \postcode{90007}
}

\author{Tongjia Yu}
\email{tongjia2428@columbia.edu}
\affiliation{%
  \institution{Columbia University}
  \city{New York}
  \state{New York}
  \country{USA}
  \postcode{10027}
}

\author{Sneha Karki}
\email{snehak@mtu.edu}
\affiliation{%
  \institution{Michigan Technological University}
  \city{Houghton}
  \state{Michigan}
  \country{USA}
  \postcode{49931}
}
\author{Tyler Zetty}
\email{tjzetty@mtu.edu}
\affiliation{%
  \institution{Michigan Technological University}
  \city{Houghton}
  \state{Michigan}
  \country{USA}
  \postcode{49931}
}

\author{Israat Haque}
\email{israat@dal.ca}
\affiliation{%
  \institution{Dalhousie University}
  \city{Halifax}
  \state{Nova Scotia}
  \country{Canada}
  \postcode{B3H 4R2}
}

\author{Shan Zhou}
\email{shanzhou@mtu.edu}
\affiliation{%
  \institution{Michigan Technological University}
  \city{Houghton}
  \state{Michigan}
  \country{USA}
  \postcode{49931}
}

\author{Dukka Kc}
\email{dbkc@mtu.edu}
\affiliation{%
  \institution{Michigan Technological University}
  \city{Houghton}
  \state{Michigan}
  \country{USA}
  \postcode{49931}
}

\author{Ian Stockwell}
\email{istock1@umbc.edu}
\affiliation{%
  \institution{University of Maryland, Baltimore County}
  \city{Baltimore}
  \state{Maryland}
  \country{USA}
  \postcode{21250}
}

\author{Albert Bifet}
\email{abifet@waikato.ac.nz}
\affiliation{%
  \institution{University of Waikato}
  \country{New Zealan}
}

\author{Wenbin Zhang}
\email{wenbinzh@mtu.edu}
\affiliation{%
  \institution{Michigan Technological University}
  \city{Houghton}
  \state{Michigan}
  \country{USA}
  \postcode{49931}
}

\renewcommand{\shortauthors}{Zichong Wang et al.}


\begin{abstract}

Bias in machine learning has rightly received significant attention over the last decade. However, most fair machine learning (fair-ML) work to address bias in decision-making systems has focused solely on the offline setting. Despite the wide prevalence of online systems in the real world, work on identifying and correcting bias in the online setting is severely lacking. The unique challenges of the online environment make addressing bias more difficult than in the offline setting. First, Streaming Machine Learning (SML) algorithms must deal with the constantly evolving real-time data stream. Second, they need to adapt to changing data distributions (concept drift) to make accurate predictions on new incoming data. Adding fairness constraints to this already complicated task is not straightforward. In this work, we focus on the challenges of achieving fairness in biased data streams while accounting for the presence of concept drift, accessing one sample at a time. We present \textit{Fair Sampling over Stream ($FS^2$)}, a novel fair rebalancing approach capable of being integrated with SML classification algorithms. Furthermore, we devise the first unified performance-fairness metric, \textit{Fairness Bonded Utility (FBU)}, to evaluate and compare the trade-off between performance and fairness of different bias mitigation methods efficiently. FBU simplifies the comparison of fairness-performance trade-offs of multiple techniques through one unified and intuitive evaluation, allowing model designers to easily choose a technique. Overall, extensive evaluations show our measures surpass those of other fair online techniques previously reported in the literature.



\end{abstract}





\maketitle

\section{Introduction}

Machine learning-based decision-making systems are constantly increasing in importance due to their wide use in fields as diverse as criminal justice, job application screening, loan decisions, and resource allocation. However, the growing adoption of automated decision-making systems has led to heightened scrutiny of these models' issues of fairness and accountability ~\cite{datta2014automated, sweeney2013discrimination, zliobaite2015survey,zhang2022kis,zhang2022fairness,zhang2023fairness}. The example of Amazon scrapping an automated hiring tool after it was found to be biased against women ~\cite{lee2019algorithmic} is one of many ~\cite{prates2020assessing, crawford2016artificial}. 
The harmful impacts of such biases makes it imperative to develop fairness-conscious classifiers that lead to accurate predictions free of discrimination against marginalized subgroups in society. While various strategies have been proposed for fair decision-making systems,  for example, to identify and eliminate discrimination~\cite{hajian2016algorithmic, wang2020visual,zhang2022longitudinal,zhang2021fair}, and ameliorate the intrinsic bias in historical data ~\cite{zhang2021farf,zhang2020deep,wang2023towards,zhang2023indiv}, they ignore a crucial space that is all around us in the real-world: the online setting.


In the real world, data is generated in real-time, and its characteristics and distribution may vary over time. Hence, it is necessary to design non-discriminatory decision systems that consider the unique challenges of the online setting. 
However, most work in fair-ML approaches fairness as a static issue, and makes the assumption that all data can be afforded for multiple scans and that the underlying population characteristics do not evolve over time ~\cite{hajian2016algorithmic, verma2018fairness,tang2020using,zhang2018content,zafar2019fairness}. Therefore they don't address the challenges faced in the online setting.
The first significant challenge is \textbf{handling the constantly produced, real-time data stream}. In other words, the data is infinite. Since no memory can load infinite data, fairness-aware learning for the online setting should be able to process each incoming instance ``on arrival,'' without requiring storage and reprocessing. Most fair-ML, however, assumes the entire dataset is available to be scanned multiple times. The second significant challenge is that \textbf{the target concepts may also evolve over time}. In other words, the instance at time $t + 1$ may have been generated by a different function than the one that generated the instance at time $t$. For example, consider the following medical diagnostics scenario. In early 2020, most patients presenting with symptoms such as fever, cough, etc would likely have received a diagnosis of the seasonal flu. However, over time we became aware of Covid-19, and while in the beginning Covid-19 would have represented a minority of the cases as compared to flu, as time passed this new disease became the majority class. What was once the minority class became the majority class, and therefore the class we would have to balance in our dataset changed over time. Addressing bias as well as a changing distribution concurrently is difficult. The third significant challenge is \textbf{an intuitive measure that is capable of quantifying the trade-off between fairness and performance}. So far, fairness evaluations have primarily reported fairness improvement and accuracy loss as two separate metrics without a reliable method for jointly measuring the inherent trade-off between them. However, in practical real-world applications, business analytics for example require a single metric jointly for both~\cite{zafar2019fairness}. Therefore, a unified and efficient measurement system is essential for clear comprehension of the trade-off between fairness and accuracy of the model.

To address the aforementioned challenges, we present Fair Sampling over Stream ($FS^2$), a novel meta-strategy for online stream-based decision-making and Fair Bonded Utility ($FBU$), a novel unified fairness-performance trade-off measurement. 
Our approach eliminates the bias in the data stream prior to applying the SML classification algorithm and is thus model agnostic. The novel fairness improvement method takes into account continuous data streams, eliminates discrimination, and accounts for evolving data distribution. Additionally, it strikes an effective and robust balance between prediction accuracy and fairness performance. Our major contributions are: 
\begin{itemize}
    \item $FS^2$, a meta-strategy that builds on the popular SMOTE technique (only applicable to batches), and can be integrated with any data stream classifier. Unlike current approaches to fair stream learning that rely on SMOTE, $FS^2$ is able to add a fairness guarantee without performance degradation, and is not restricted to batches. 
    \item A novel fairness-performance metric that assesses the fairness-accuracy trade-off of ML bias mitigation methods in a unified and intuitive manner.
    \item Qualitative and quantitative experiments on five groups of  biased data streams show the effectiveness of the proposed unified metrics and fairness-conscious online learners in streaming environments.
\end{itemize}
The remainder of the paper is organized as follows. Section ~\ref{sec:related_work} and Section ~\ref{sec:notations} review relevant work and the theoretical background knowledge regarding discrimination-aware learning. Next, we discuss the limitations of class balancing technique in Section~\ref{EFFECTS OF CLASS BALANCING ON FAIRNESS}. We present $FS^2$ and bias-performance measurement metric in Section ~\ref{sec:methodology}. Section ~\ref{sec:experiment} analyzes the experimental results. Finally, we conclude the paper in Section~\ref{sec:conclusion}. 
\vspace{-0.3cm}
\section{Related Work}
\label{sec:related_work}


\subsection{Fairness-aware Learning}

A number of approaches have been proposed to address the problem of bias and discrimination in machine learning, and can be broadly grouped under three categories depending on the part of the pipeline fairness interventions are applied to: pre-, in-, or post-processing. i) \textit{Pre-processing techniques} focus on mitigating and correcting bias in the data used for training the model, and therefore are model-independent. The most
popular ones include massaging~\cite{kamiran2009classifying} and reweighting~\cite {calders2009building}, which have also been extended for addressing bias in the input data before updating the online model in~\cite{iosifidis2019fairness}. ii) \textit{In-processing techniques} transform an algorithm to improve fairness, typically by embedding fairness in the objective function by means of regularization or other constraints. \cite{kamiran2010discrimination} is among the first line of works in this category by integrating discrimination into the splitting criteria for fair tree induction. More recently,~\cite{zhang2019faht} further improves this splitting criteria and applies it in the online settings. iii) \textit{Post-processing techniques} are based on either adapting the decision boundary or simply modifying the output of a model (e.g. the prediction labels). With supplemental prediction thresholds in place, \cite{hardt2016equality} works against discrimination while in \cite{calders2010three} the decision boundary of AdaBoost is adjusted with regards to fairness. We would like to stress on the fact that transferring such approaches to an online setup is not as simple since boundary/prediction might evolve due to the non-stationary spread in an online setting.

\vspace{-0.3cm}
\subsection{Stream Learning}

The main challenges of learning in a stream setting are concept drift, where the joint data distribution changes over time and class imbalance~\cite{krawczyk2017ensemble,aggarwal2007data,gama2014survey}. To address the first challenge, learning methods need to adapt to changes incrementally by incorporating new information into the model~\cite{wang2013learning,wang2014resampling,read2019error}, and by forgetting outdated information~\cite{ghazikhani2014online, ghazikhani2013ensemble,zhang2017hybrid}. For the second challenge, various sampling methods have been proposed including the most representative (C-SMOTE) \cite{bernardo2020c} to address class imbalance in a continuous data stream while ensuring the model can adapt to shifts in the data distribution. However, these techniques focus entirely on accuracy without considering fairness.
\vspace{-0.3cm}
\subsection{Fairness-aware Stream Learning} 
Despite the wide prevalence of the stream learning setting as well as bias and discrimination within the real world, surprisingly little work has explored this setting so far. In addition to the aforementioned fair online methods, \cite{bechavod2020metric} focuses on individual level and requires the existence of an auditor to detect fairness violations, while Online SMOTE Boost~\cite{wang2016online} is made cost-sensitive by adding various parameters of the Poisson distribution for different classes to address bias. In addition, FABBOO~\cite{iosifidis2021online} adjusts the training distribution on-the-fly, considering both the imbalanced nature of the stream and the model's discriminatory behavior as evaluated from past data. One common drawback to all of these methods is that their balancing approaches are driven by the SMOTE which could further exacerbate bias (c.f., Section~\ref{EFFECTS OF CLASS BALANCING ON FAIRNESS}). Our method addresses this drawback by performing an additional clustering for fairness guarantee.

\section{Basic Concepts}
\label{sec:notations}

Let $D$ be a sequence of instances $d_1, d_2, \dots, d_n$ that arrive over time at timepoints $t_1, t_2, \dots, t_n$, where each instance $d \in \mathbb{R}$. Similarly, let $Y$ be a sequence of corresponding class labels, such that for each $d \in D$ there exists a corresponding class label $y \in Y$. 
We assume each data instance in the continuous stream has the form $d = {\{A, S, Y\}}$, where $A$ is a set of attributes, $S$ denotes the sensitive attribute (e.g. gender/race), and $Y$ is the true class label. 
In fair-ML, a sensitive attribute is a characteristic legally protected from discrimination (e.g., race, color, gender, etc). Typically, one sensitive attribute class has been historically marginalized, such as people of color or women, and another that has not. We refer to the former as the \textit{unprivileged group} and the latter as the \textit{privileged group}. Without loss of generality, we assume a binary classifier, $f(): D \rightarrow	y$, such that $y \in \{-1, +1\}$. The class label predicted by this classifier shall be denoted by $\hat{Y}$. 

Since we explore the \textit{online learning environment}, each new data instance from the continuous data stream shall be processed one by one. A characteristic of the online learning setting is that the true class label of each instance is revealed to the learner before the arrival of the next instance. Therefore, the way things proceed in the online setting is as follows. For each new data instance $d$ arriving at time point $t$, a class label is predicted using $f_{t-1}$, i.e. the classifier from time point $t-1$. The true class label of data point $d$ is revealed to the online learner before the arrival of the next data instance. The model is then updated using this true class label to get $f_t$. This sequence of events is referred to as the \textit{first-test-then-train} or \textit{prequential evaluation} setup~\cite{gama2010knowledge}. 

Another characteristic typical to the online setting is the idea of \textit{concept drift}. Concept drift refers to the changes in underlying distribution of the continuous data stream over time. In other words, it denotes the changes in the joint distribution. Mathematically, $P_{t_a}(D, Y) \neq P_{t_b}(D, Y)$ for two different timepoints $t_a$ and $t_b$. Such changes often render the current classifier ineffective, since it is unable to deal with the changes in the data distribution, thereby necessitating a model update. In this work we focus on the scenario where the continuous data stream is also imbalanced, i.e. one class is much more represented than the other. Machine learning models typically neglect the minority class in such situations to prevent overfitting and loss of generalization ~\cite{weiss2004mining}. Our technique does not require the minority class to be predefined in advance, nor does it assume that the same class will always be the minority class. In other words, we allow for the possibility that role of the minority class may alternate between the two classes. We analyze the extent of the imbalance in data by computing the imbalance ratio (IR) with  Equation~\ref{equ:Imbalance Ratio}:
\begin{equation}
\label{equ:Imbalance Ratio}
    IR = \frac{S_{t_{min}}}{S_{t_{maj}} + S_{t_{min}}}
\end{equation}
where $S_{t_{min}}$ is the minority class and $S_{t_{maj}}$ is the majority class in the data stream at time $t$.






The traditional approach to fairness-aware classification in the offline setting seeks to learn a mapping, $F:(A, S, Y) \Rightarrow Y$, where the goal is to assign a class label to each data instance accurately without discriminating against the unprivileged group. The desirable class label (e.g. receiving a loan in loan decisions; receiving bail in bail decisions) is known as the \textit{`favorable' label}, and the other, undesirable class label (e.g., being denied a loan or bail) is called the \textit{`unfavorable' label}. We focus on parity-based approaches that compare a fair-ML model's behavior towards the unprivileged and privileged groups. The two measures we employ are statistical parity ~\cite{kamiran2012data}, and equal opportunity ~\cite{hardt2016equality}. We extend them to the online setting. 

Statistical parity quantifies the disparity between the probability of the privileged and unprivileged groups receiving a benefit. Cumulative Statistical Parity Difference ($CSPD$) in Equation \ref{equ:Statistical Parity Difference} represents the cumulative difference in statistical parity between the privileged and unprivileged groups over time. $\lambda \in [0, 1]$ is a decay factor used to regulate the amount of influence of previous time points. In other words, the larger $\lambda$, the higher the contribution of historical information. It is specified by the user.
\begin{equation}
\label{equ:Statistical Parity Difference}
    CSPD_t = (1 - \lambda ) \times (\frac{\sum_{i=1}^{t} (F_i(d_i) \land Y = 1 | S = \overline{s} )}{\sum_{i=1}^{t} ( d_i | S = \overline{s} } - \frac{\sum_{i=1}^{t} (F_i(d_i) \land Y = 1 | S = s )}{\sum_{i=1}^{t} ( d_i | S = s ) } )+ \lambda \times CSPD_{t-1}
\end{equation}


However, statistical parity may result in the phenomenon of reverse discrimination in certain conditions (i.e. the scenario when data instances that are not deserving of a positive assignment receive one incorrectly), since it does not consider real class labels. Therefore, we also consider Equal Opportunity, which measures the difference in True Positive Rates (TPR) between the privileged and unprivileged groups. 
Cumulative Equal Opportunity Difference $CEOD$ in Equation \ref{equ:Equal_Opportunity_Difference} is the cumulative difference in equal opportunity between the privileged and unprivileged groups over time. 
\begin{equation}
\label{equ:Equal_Opportunity_Difference}
    CEOD_t = (1-\lambda) \times (\frac{\sum_{i=1}^{t} (F_i(d_i) \land Y = 1 | S = \overline{s}, Y = 1)}{\sum_{i=1}^{t} ( d_i | S = \overline{s}, Y = 1) } - \frac{\sum_{i=1}^{t} (F_i(d_i) \land Y = 1 | S = s, Y = 1)}{\sum_{i=1}^{t} ( d_i | S = s, Y = 1) }) + \lambda \times CEOD_{t-1}
\end{equation}
where $\land$ is the logical symbol and $\lambda$ is used for correction in the early stages of the stream (division by 0).

\section{EFFECTS OF CLASS BALANCING ON FAIRNESS}
\label{EFFECTS OF CLASS BALANCING ON FAIRNESS}

Inequality in class distribution is a hallmark of biased data. As the instances in the underprivileged group are not as prevalent as the privileged group, the model may overlook patterns in the underprivileged group classification. Since machine learning is data-driven, it is easy for a model trained on a biased dataset to assign favorable labels to the privileged subgroup and unfavorable labels to the unprivileged subgroup in pursuit of higher accuracy. To address unequal representation, class balancing techniques such as oversampling and undersampling are used. Oversampling increases the minority class representation by creating synthetic instances, while undersampling balances representation by reducing majority class instances. We examine the drawbacks of these techniques and how they affect the bias and fairness in a model's output in detail. While certain limitations of popular class-balancing techniques are well known (e.g., oversampling may lead to overfitting if samples from a few classes are replicated multiple times~\cite{yan2020fair,douzas2018improving,chakraborty2021bias}; undersampling may lead to hidden information being ignored due to the samples that were removed~\cite{sharma2017pros,zhang2017hybrid,chawla2002smote}), some other drawbacks are not as well studied, \begin{wrapfigure}{r}{0.4\textwidth}
	\centering
	\includegraphics[width=0.4\textwidth]{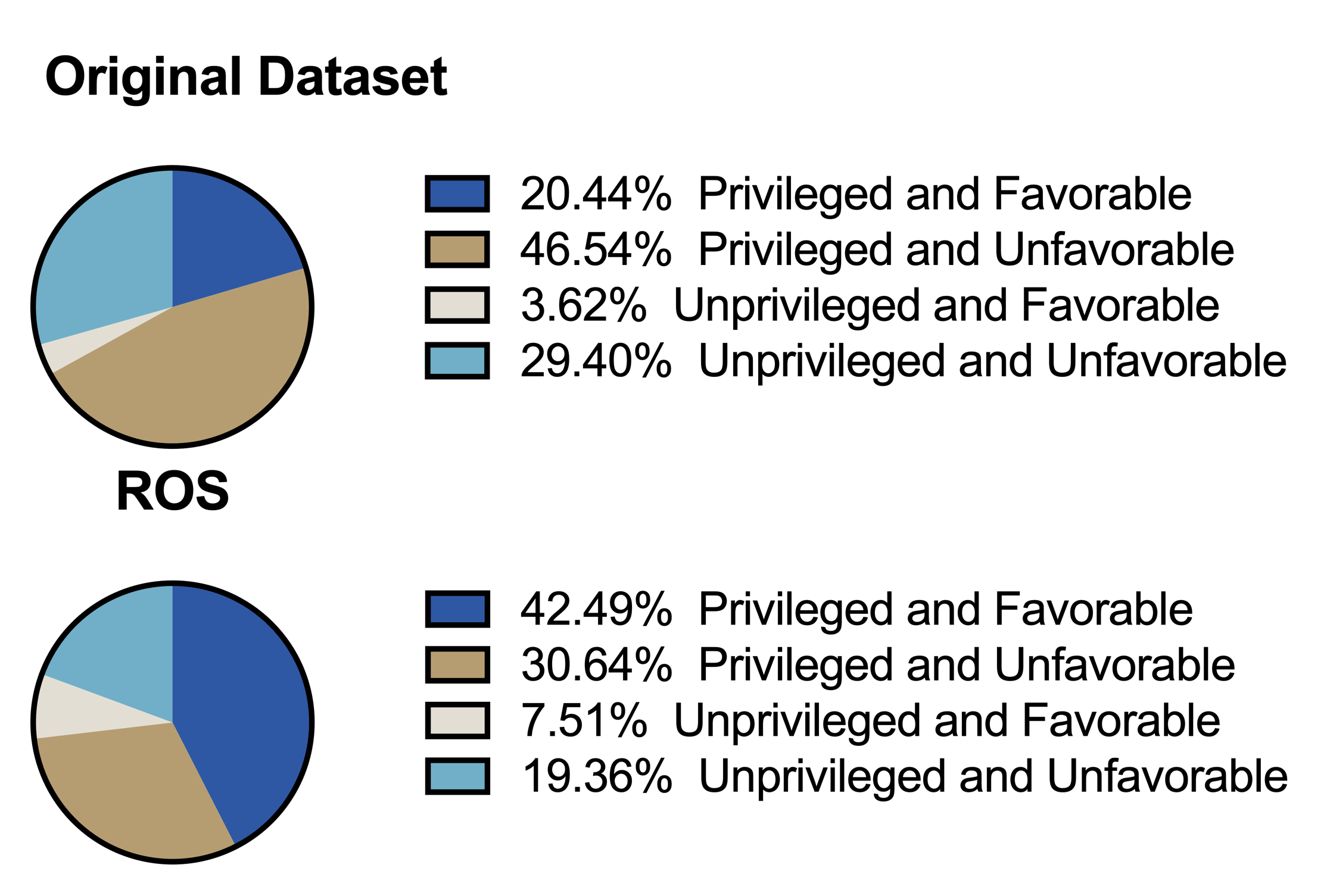}\vspace{-0.3cm}
	\caption{The change of distribution bias in the Bank Marketing dataset before and after class balancing.}
	\label{fig:limit}
\end{wrapfigure}particularly in terms of fairness research when alleviating bias.

Traditional class-balancing techniques make the minority class samples statistically equal to the majority class samples. However, since they aim to solve the imbalance only in terms of the target variable (i.e. the class label), this process is blind to the inherent properties of the dataset. As a result, the bias in the dataset may actually increase after balancing the class labels, causing the model's fairness to decrease further. This is especially problematic in the case of multiple privileged and unprivileged subgroups. Not all privileged subgroups are privileged in the same way and/or to the same extent, nor are all unprivileged subgroups unprivileged in the same way, and/or to the same extent. Undersampling can lead to a disproportional increase in the number of instances of the more favorable privileged group. On the other hand, oversampling may result in a disproportionate increase in the number of instances of the more favored unprivileged group because they may be a bigger proportion of the minority class, leaving the worst favored unprivileged group severely underrepresented.  After extensive experiments we see in Fig.~\ref{fig:limit} that using common class balancing techniques can indeed lead to more distribution bias in the dataset. In Fig.~\ref{fig:limit} the original Bank Marketing dataset, only 24\% of the samples are positive (subscribe) while the rest are negative (non-subscribe). After class balancing with Random Over Sampling (ROS), 11,788 more positive samples were added, of which only 15\% were unprivileged subgroup. This increases bias in the distribution, as more privileged subgroup samples were added during class balancing. In other words, the representation of unprivileged favorable groups in the data set declines further. Fig.~\ref{fig:smote} represents a limitation of another common class balancing technique, SMOTE. As the Figure~\ref{fig:smote} shows, the minority class samples synthesized by SMOTE may belong to the majority class in the feature space. Meanwhile, random selection of samples for synthesizing instances can lead to further in-class bias in the dataset, especially in datasets with multiple sensitive attributes.

\section{Methodology}
\label{sec:methodology}

In this section, first we introduce our proposed performance-fairness metric, Fairness Bonded Utility (FBU). FBU is the first metric to unify the fairness-performance trade-off into one intuitive and versatile metric capable of comparing any combination of fairness and performance metrics. Then we present Fair Sampling over Stream ($FS^2$), a fair rebalancing technique for a biased data stream capable of being integrated with any stream-learning classification technique.

\subsection{Fairness Bonded Utility}
\label{measurement}

A long-standing problem in fair-ML has been the trade-off dance that occurs between model performance and fairness. \begin{wrapfigure}{r}{0.4\textwidth}
	\centering
	\includegraphics[width=0.26\textwidth]{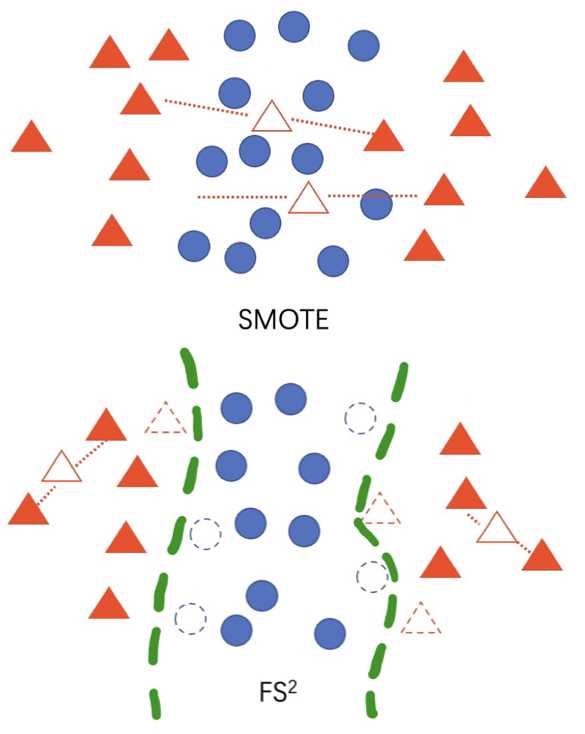}\vspace{-0.3cm}
	\caption{Illustration of oversampling by SMOTE and $FS^2$. Circle and Triangular nodes symbolize samples with majority and minority labels, respectively. The hollow nodes are samples created based on the closest neighbors of minority samples within each cluster.}
	\label{fig:smote}
\end{wrapfigure}Typically, a model's performance degrades as fairness guarantees become stronger. This complicates the decision about which fairness metric to use for which application since analyzing this trade-off between two different numbers (performance and fairness) is not always intuitive. Fairness Bonded Utility (FBU) solves this problem by assigning each fairness technique into one of five intuitive ``effectiveness'' levels ranging from ``win-win'' (fairness and performance both improve) to ``lose-lose'' (both degrade). At a high level, FBU does this by measuring how the performance of a model varies with different fairness techniques in a two-dimensional coordinate space and computing a trade-off baseline that categorizes the various fairness techniques into effectiveness levels. FBU can drastically simplify the agonizing decision-making process of which fairness technique to use for which application scenario.

\begin{figure}[!htbp]
	\centering
	\includegraphics[width=0.55\textwidth, height=0.30\textwidth]{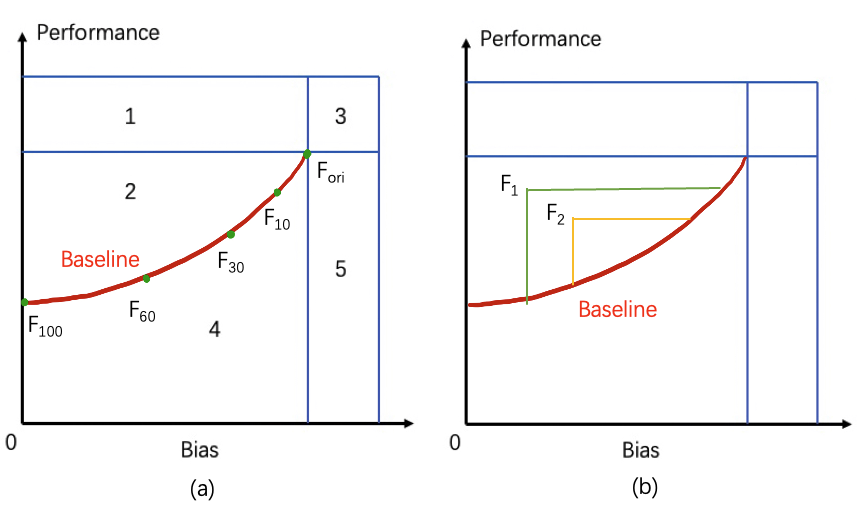}
	\caption{The FBU fairness-accuracy trade-off baseline is depicted by the original trade-off point ($F_{ori}$) and the points generated by the pseudo models ($F_{10}$,$\dots$, $F_{100}$). A bias reduction method is considered effective if it shows a superior trade-off compared to the FBU baseline (i.e., it lies above the red line).}
	\label{fig:fbu}
\end{figure}

Before going into details, we first construct the environment for the FBU to measure the performance of different fairness techniques. Mathematically, it is a two-dimensional coordinate system (see Figure \ref{fig:fbu}). The $x$-axis represents the model's bias, and the $y$-axis depicts the model's performance. Our technique is versatile and can be used with any fairness metrics to measure bias (e.g., CSPD, CEOD, etc.) and any performance metrics (e.g., accuracy, f1-score, etc.) of the model designer's choosing. Next, we outline the creation of the trade-off baseline and the evaluation of the outcomes.

\textbf{Creating a Trade-off Baseline}: 
The idea behind the trade-off baseline in FBU is centered around the zero-normalization principle presented by Speicher et al. ~\cite{sun2020automatic}. It posits that bias is minimized when every individual receives the same label. In other words, bias would be 0 when everyone receives the same assignment as the output, although this may lead to decreased performance, and vice versa. 

We use this principle to create multiple pseudo-models, $F_p$, where $p \in \{10\%, 20\%, \dots, 100\%\}$. In each pseudo-model, we substitute a proportion $p$ of the original model's predictions with the same label. We vary the proportion of substituted predictions in each pseudo model, with $p$ ranging from 10\% to 100\%. In other words, we will replace 10\% of the original model's predictions with the same label in $F_{10}$. In $F_{100}$ we replace 100\% of the original model's predictions with the same label. As the proportion of predictions substituted increases, the model's fairness will increase but its accuracy will decrease. 

Now we use the original model (with no fairness technique) and the created pseudo models to form the trade-off baseline. We derive the first point for the trade-off baseline from the (performance, bias) coordinate of the original model (point $F_{ori}$ in Fig. ~\ref{fig:fbu}(a)). Then we plot the (performance, bias) coordinates of the ten pseudo models (i.e., $F_{10}, F_{30}, F_{60}$, etc. in Fig. ~\ref{fig:fbu}(a)). Finally we connect them all together to form the trade-off baseline.

\textbf{Five effectiveness levels}: The trade-off baseline categorizes the different bias mitigation techniques being compared into five intuitive effectiveness levels, each with a different bias-performance trade-off. 
The first level, region 1 in Fig. ~\ref{fig:fbu}(a), is a ``win-win'': A technique belongs in this category if it improves both model performance and fairness relative to the trade-off baseline. 
Region 2 in Fig. ~\ref{fig:fbu}(a) denotes a ``good'' trade-off: a technique in region 2 enhances either model performance or fairness compared to the trade-off baseline, making it overall better than the trade-off baseline.
If a technique improves model performance but leads to a decrease in fairness, it is considered an ``inverted'' trade-off and falls into region 3 (Fig. ~\ref{fig:fbu}(a)). 
A technique in region 4 in Fig. ~\ref{fig:fbu}(a) is a ``poor'' trade-off: it leads to a decline in either model performance or fairness compared to the baseline, making it overall worse than the trade-off baseline. 
A region 5 technique (in Fig. ~\ref{fig:fbu}(a)) represents a ``lose-lose'' trade-off since it decreases both model performance and fairness compared to the original model. 

\textbf{Quantitative Evaluation of Trade-offs}: The ``win-win'' (region 1), ``lose-lose'' (region 5), and ``poor'' (region 4) trade-off regions offer clear indications of the effectiveness of the bias mitigation technique. Therefore, we concentrate on quantifying the trade-off quality of bias mitigation techniques in the ``good'' trade-off region to enable a more detailed comparison. Specifically, we will measure the strengths and weaknesses of different bias mitigation methods that fall in the ``good'' region (region 2) based on the area enclosed by the bias-performance points and baseline (see Fig. \ref{fig:fbu}(b)). Techniques with larger areas have better, more desirable bias-performance trade-offs. Using area as a measure of trade-off, instead of other criteria such as distance from the trade-off baseline, guarantees a fair comparison even in cases where the trade-off baseline is curved.

The final output of FBU for a technique will be five percentages: one percentage for each region representing how many cases for that technique fell in that region. The number of cases per region is computed by: $n_r\ x\ n_t\ x\ n_f\ x\ p_m $, where $n_r$ is the number of run times, $n_t$ number of techniques compared against, $n_f$ number of fairness metrics used, and $p_m$ is the number of performance metrics used. 



\subsection{Fair Sampling over Stream}
\label{sec:FCSMOTE}
There are two fundamental challenges facing Fair Sampling over Stream ($FS^2$): the first is the lack of the entire dataset in the rebalancing phase. The second is synthesizing fair data. We first discuss the online monitoring of the continuous data stream and model updates. Then we elaborate on the challenge of synthesizing fair data. We follow with the $FS^2$ algorithm, and end the section with explaining $FS^2$ classification. 

\subsubsection{Online Monitoring and Model Update}
\label{sec:OnlineMonitoringandModelUpdate}

In continuous data streams, the status of minority and majority classes can evolve. In other words, a class currently identified as a minority may turn into the majority class, or vice versa ~\cite{wang2013learning}. Additionally, approaches focusing solely on fairness in recent outcomes (i.e., in the immediate future) are inadequate in combating discrimination over time. Although individual instances of discrimination may seem minor when evaluated alone, they can accumulate and result in substantial bias in the long term. Hence, monitoring the subgroup distribution in the data stream over time is vital for both the efficiency and fairness of the model. However, the primary challenge faced by $FS^2$ is its inability to consistently monitor data to address both concept drift and fairness concerns. In streaming machine learning, we assume the data stream to be infinite. Firstly, no memory can physically store infinite data; secondly, this would be against the stream paradigm approach. Therefore, storing every new sample in memory is impossible until the data stream ends. 

To address this problem, we employ a different approach from previous non-stationary studies ~\cite{chen2012online, gomes2017adaptive}. $FS^2$ employs ADWIN ~\cite{inproceedings} to detect changes in both accuracy and fairness, reflecting both concept and fairness drifts. In other words, drift is detected when either fairness or the data distribution evolves. Specifically, we will use ADWIN to save the most recently seen samples and apply the synthesis algorithm (explained below in Section \ref{Synthetic fair data}) using the samples stored in ADWIN. ADWIN maintains a variable-length window of recently seen items and can automatically detect and adapt the window size to the current rate of change in the data. As shown in Equation ~\ref{equ:levelError}, ADWIN employs a threshold known as $\delta$ (i.e., performance threshold and fairness threshold) to configure the error to have two levels automatically. These levels are known as the warning level and the change level.
\begin{equation}
\label{equ:levelError}
    levelError = \log(\frac{2 \times \log n }{\delta})
\end{equation}
We identify the warning level by multiplying $\delta$ by 10, and $\delta$ itself helps determine the change level. Since $\delta$ is in the denominator, multiplying it by 10 will result in a lower value than the one obtained by using $\delta$ alone. Therefore, we will reach the warning level before the change level and can detect changes early to take appropriate action. The width of the window at any given moment is represented by $n$. 


ADWIN monitors the error that occurs over the data in the window. If the error reaches a threshold of the warning level, ADWIN assumes that either a concept drift or fairness drift is starting to occur, and it starts collecting new samples in a new window. If the error exceeds the change level, ADWIN assumes either a concept drift or a fairness drift has occurred, and it substitutes the old window with the new one. 


\subsubsection{Synthetic fair data}
\label{Synthetic fair data}

Here we propose a new strategy for synthesizing fair data to address the challenges posed by class balancing and its impact on model fairness. Our proposed approach eliminates the in-class and between-class imbalance in the oversampling part. Unlike traditional class balancing techniques such as SMOTE, our method identifies the true minority classes in the dataset (see Figure \ref{fig:smote}), and adds fairness constraints to the generation process for synthetic samples.
At a high level, rather than directly running SMOTE, our approach first determines similar samples via clustering, then filters the clusters using silhouette score to ensure a good clustering. Finally, it uses SMOTE within each cluster to generate synthetic minority samples. Notably, our approach not only enhances the fairness of the balancing method but also preserves the accuracy of the prediction. Below we discuss our approach in detail. 

First, we use a clustering algorithm to identify homogeneous subgroups of feature similarity in the feature space. In our experiments, we use the KNN algorithm. Note that any clustering algorithm can be applied, but different clustering algorithms may have different clustering results and time complexity. By partitioning the dataset according to the selected number of clustering families $M$, samples with similarities in the feature space will be classified into the same clusters. We will measure the quality of clustering by silhouette analysis. The silhouette score measures how similar an object is to its own cluster compared to others, and the value is between [$-1$, $1$]. A value of $1$ indicates perfect separation, and a value of $-1$ suggests that the object is in the wrong cluster. Values between $0$ and $1$ show the degree of separation, with values closer to $1$ indicating a stronger separation.  

After determining the optimal number of clusters $M$, the algorithm performs filtering detection for each cluster. The algorithm will calculate the silhouette score of each sample and then remove samples with the lowest 20\% silhouette score from the dataset. The intuition for this is as follows. Let's imagine a realistic data distribution situation where a sample of a minority group exists near the cluster boundaries. In this case, we can assume that SMOTE selects the minority group sample near the clustering boundary as the template. Then the simulated data generated based on this sample will also be close to the clustering boundary. In addition, there is a risk that the generated simulated data are even closer to the clustering boundary than the original data. This increases the number of samples near the cluster boundaries, thus blurring the boundaries between different clusters, making it difficult to separate these samples in the feature space and a higher probability of being assigned to incorrect clusters. As a result, new instances generated based on these samples may further increase the distribution bias. 


After the filtering is complete, the algorithm checks each cluster. Specifically, the algorithm detects the proportion of minority class samples in each cluster to determine its synthetic weight, which is directly related to the proportion of minority samples in the cluster. If the percentage of minority samples in a cluster is higher, the more representative the cluster is for minority samples. In other words, clusters with a high percentage of minority class samples are assigned higher synthetic weights. For each cluster, the synthetic weight is calculated according to Equation ~\ref{equ:calweight}:



\begin{equation}
\label{equ:calweight}
    Cluster Weight = \begin{cases}
        W_i = \frac{N_i}{N_{Total}} \\
        W_1 + W_2 + \dots + W_n = 1
    \end{cases}
\end{equation}
where $N_i$ is the number of minority class samples in each cluster, $N_{Total}$ is the total number of minority class samples, and $W_i$ is the synthetic weight of cluster $i$. The sum of all weights is one, and the number of samples selected for each cluster is calculated according to Equation~\ref{equ:eachclusternumber}: 
\begin{equation}
\label{equ:eachclusternumber}
    G_i = N_{num} \times W_i
\end{equation}
where $N_{num}$ is the total number of samples to be synthesized. After calculating the number of samples selected for each cluster, the model will randomly select samples in the clusters, and a new minority sample is generated according to the SMOTE~\cite{chawla2002smote} generation algorithm based on the k-nearest neighbors of the minority samples. 

\subsubsection{$FS^2$ Algorithm}
\label{$FS^2$}

The algorithm for $FS^2$ is outlined in Algorithm~\ref{alg:algorithm1} using pseudocode. 

The $FS^2$ algorithm utilizes two sliding windows of samples, $W$ and $W_{label}$ (line 1). $W$ contains all samples related to the current concept. $W_{label}$ is an array indicating whether the corresponding samples in $W$ belong to the privileged favorable subgroup, privileged unfavorable subgroup, unprivileged favorable subgroup, or unprivileged unfavorable subgroup. The $FS^2$ algorithm utilizes nine counters (lines 2-4): $C_0$, $C_1$, $C_2$ and $C_3$ are used to count the number of instances in each subgroup, while $\overline{C_0}$, $\overline{C_1}$, $\overline{C_2}$ and $\overline{C_3}$ count the number of instances of different subgroups generated by the synthesis algorithm, respectively. The $C_{gen}$ counter for each sample $d_i$ in the sliding window $W$ keeps track of the number of times $d_i$ is used to generate synthetic samples. 

$FS^2$ tracks instances of different classes. As we mention above, after new samples are continuously collected, the number of samples of different classes in $W$ can deviate significantly. In this case, the representation of different subgroups in $W$ may change, $FS^2$ will synthesize new samples to balance the representation of different subgroups. In line 5, the variable $adwin$ is a change detector that ensures the two windows ($W$, $W_{label}$) remain consistent with any concept drift by using the class value of the incoming sample. $adwin$ is used later for adapting the model to concept and/or fairness drift as needed.  In line 6 the variables $balR, fairR$ are initialized to 0.  

For each new instance $d_i$ (line 7), we first train the pipeline learner $l$ using the new sample (line 8). After this process, the new sample $d_i$ is saved in the sliding window $W$ (line 9). On line 10, the function $updateWindows$ adds a new bit into $W_{label}$ depending on the new sample's class label $Y$ (i.e., 0 if $Y$ = 0). On line 11, the function $updateCounters$ increments the relative counter $C_0$, $C_1$, $C_2$, or $C_3$ depending on the sample's class label $Y$ and sensitive attribute $S$. 

After that, the $adwin$ is updated with the sensitive attribute $S$ and class value $Y$ of the incoming sample $d_i$ (line 12) and the $checkDrift$ function uses $adwin$ to check for any concept or fairness drift (line 13). The function adapts by reducing $W$ in accordance with the window maintained by $adwin$ to the occurrence of concept drift or fairness drift. Additionally, it updates $W_{label}$ and the counters $C_0$, $C_1$, $C_2$ and $C_3$ based on this reduction. Furthermore, if the instances removed from $W$ have been used to create synthetic samples, the function also updates the counter of instances generated for that class ($\overline{C_0}$, $\overline{C_1}$, $\overline{C_2}$ or $\overline{C_3}$). Line 14 determines the actual representation of each subgroup and then saves the associated window and counters into $W_{syn}$, $C_{syn}$, $\overline{C_{syn}}$. 

Then we evaluate whether the instances in $W_{synthetic}$ satisfy or surpass the hyperparameter $minSize$ set by the user (line 15). $minSize$ denotes the minimum number of data samples already seen before rebalancing can occur. For example, if $minSize = 5$, that means we must have seen at least 5 data samples before rebalancing can occur. 
If both criteria are met, the rebalancing stage begins. If not, the algorithm waits for another data sample to arrive. The imbalance ratio between the number of minority and majority instances is calculated according to Equation ~\ref{equ:Imbalance Ratio}, and Equation~\ref{equ:Statistical Parity Difference} is used to compute the statistical parity difference. The results are stored in $balR$ and $fairR$, respectively (line 16). 

Now we present the rebalancing part of the algorithm. 
If $balR$ is less than $p_1$ it means that $W$ is unbalanced, and if $fairR$ is less than the hyperparameter $f_1$, then $W$ is unfair. 
Therefore, new synthetic samples $\hat{d}_i$ are generated using the function $FairGenerate$ until $W$ is fair and balanced (line 18). In other words, the condition to stop synthesizing samples is that $balR$ equals $p_1$ and $fairR$ equals $f_1$, As described in the previous section, our method of generating samples will be fairer than the SMOTE. 

Before generating new data points, $FS^2$ will calculate the number of instances introduced for each batch subgroup. Then it will cluster the instances in sliding window $W$, identify similar individuals through the feature space, filter the boundary samples, and calculate the synthetic weights for the different clusters. $FS^2$ will randomly select samples in each cluster to synthesize new instances $d_i$ and update the counter $S_{gen}$ until the synthesis number of that cluster is reached. In addition, to avoid overfitting, we will keep the one-time principle, meaning that a sample will be used only once to synthesize instances during the rebalancing phase. The algorithm violates the one-time principle only in one case: when the number of synthesized instances needed to rebalance $W$ is larger than the number of samples. Only in this case will the function reuse the same samples multiple times during the same rebalancing phase. Finally, we will use the new synthesized instances $d_i$ to train the learner $l$ (line 19) and calculate the new $balR$ and $fairR$ (line 21).

\begin{algorithm}[!htb]
\caption{$FS^2$ Leaning Algorithm}
\label{alg:algorithm1}
\LinesNumbered
\KwIn{a discriminated data stream $D$, the pipeline learner $l$, optional sampling ratio $p_1$ and $f_1$}
$W, W_{label}\leftarrow \emptyset$\;
$C_0$, $C_1$, $C_2$, $C_3\leftarrow 0$\;
$\overline{C_0}$, $\overline{C_1}$, $\overline{C_2}$, $\overline{C_3}\leftarrow 0$\;
$C_{gen}\leftarrow \emptyset$\;
$adwin\leftarrow \emptyset$\;
$balR, fairR\leftarrow 0$\;
\For{each instance $d_i$ in $D$}
{
    $Train \ (d_i,l)$\;
    $W \leftarrow Add \ (d_i)$\;
    $UpdateWindows \ (d_i,W_{label})$\;
    $C_0, C_1, C_2, C_3 \leftarrow updateCounters \ (d_i)$\;
    $adwin \leftarrow Add \ (d_i)$\;
    $CheckDrift \ (adwin,W,W_{label},C_1,C_2,C_3,C_4,\overline{C_0}, \overline{C_1}, \overline{C_2}, \overline{C_3}, C_{gen})$\;
    $W_{syn}, C_{syn}, \overline{C}_{syn} \leftarrow CheckRep \ (C_1,C_2,C_3,C_4,\overline{C_0}, \overline{C_1}, \overline{C_2}, \overline{C_3})$\;
    \If{CheckSize(minSize, $C_{syn}$)}{
    $balR, fairR\leftarrow Calculate \ (C_1,C_2,C_3,C_4,\overline{C_0}, \overline{C_1}, \overline{C_2}, \overline{C_3})$ according to Equation~(\ref{equ:Imbalance Ratio}) and Equation~(\ref{equ:Statistical Parity Difference})\;
    \While{$f_1 \textgreater fairR \ or \ p_1 \textgreater balR$}
    {
        $\hat{d}_i \leftarrow FairGenerate( W_{syn},S_{gen})$\;
        $Train \ (\hat{d}_i,l)$\;
        $\overline{C}_{syn} += 1$\;
        $balR, fairR\leftarrow Calculate \ (C_1,C_2,C_3,C_4,\overline{C_0}, \overline{C_1}, \overline{C_2}, \overline{C_3})$ according to Equation~(\ref{equ:Imbalance Ratio}) and Equation~(\ref{equ:Statistical Parity Difference})\;
    }
    
  }
  
}
\end{algorithm}

\subsubsection{$FS^2$ Classification}
\label{sec:$FS^2$Classification}
$FS^2$ is a meta-strategy that solves class imbalance and cumulative discriminatory results in data streams. In particular, we have integrated the Adaptive Hoeffding Trees (AHT)~\cite{bifet2009adaptive} classifier. AHT is a stream-based decision tree induction algorithm that ensures the tree's adaptation to changes in the underlying data distribution by updating the tree with new instances from the stream and replacing underperforming sub-trees. Therefore, for the online setting, AHT is superior to the majority of other classifiers. $FS^2$ can be pipelined with any data stream classifier. 

\section{Experimental Evaluation}
\label{sec:experiment}


\subsection{Datasets}
\label{Datasets}
Online fairness research is hindered by the lack of access to good datasets~\cite{li2021time}. The large amount of data required and the necessity for concept drift to be present limit availability. We use three real-world datasets and one synthetic \begin{wraptable}{r}{7cm}
 	\caption{Summary of datasets used in experiments.}
	\begin{tabular}{|c|c|c|c|c|}
		\hline
		\diagbox{Dataset}{CHAR}  & Sample\# & Features\# & \makecell[c]{Sensitive \\ Attribute} \\
		\hline
		BNK  			& 52944       & 150     & Age   \\
		GMSC		    & 150,000        &29      & Age   \\
            NYPD 		& 311,367           & 16         &Gender \\
		SYN  		& 1,000,000        & 21      &Synth \\
		\hline
	\end{tabular}  
	\label{tab:dataset_info}
\end{wraptable}

dataset (see Table~(\ref{tab:dataset_info} for details). i) Bank Telemarketing Data (BNK)~\cite{MORO201422}: The objective of the classification task here is to determine whether a client will subscribe to a term deposit. ii) Give Me Some Credit Dataset (GMSC) ~\cite{https://www.kaggle.com/c/givemesomecredit/overview_2021}: The objective here is to determine whether to approve a loan application. iii) NYPD~\cite{sargent2017office}: We use this dataset to predict whether a suspect was convicted of a felony or not. iv) Synthetic dataset: we followed the initialization process by the authors~\cite{iosifidis2019fairness}, which involves generating each attribute as a different Gaussian distribution. To simulate class imbalance and concept drift, we introduced these elements into the data stream by shifting the mean of average of each Gaussian distribution.



\subsection{Baselines and Metrics}
\label{Baseline}

To evaluate the performance of our method, we compare $FS^2$ against four state-of-the-art online fairness streaming methods based on four metrics. The four baselines are: i) Fairness-aware Hoeffding Tree~\cite{zhang2019faht} (FAHT): FAHT is a fairness-focused variation of the Hoeffding tree algorithm that considers both performance and fairness when selecting split attributes. ii) Online Smooth Boosting~\cite{wang2016online} (OSBoost): OSBoost is made cost-sensitive by adding various parameters of the Poisson distribution for different classes. iii) Online fairness and class imbalance-aware boosting~\cite{iosifidis2021online} (FABBOO): FABOO considers long-term class imbalance to fight class imbalance and discrimination from a boosting approach. iv) Continuous SMOTE~\cite{bernardo2020c} (C-SMOTE): C-SMOTE is an approach that addresses class imbalance by resampling the minority class using a sliding window, but does not consider fairness.

In terms of the four metrics, two performance metrics and two fairness metrics are employed. i) Balanced Accuracy: It is the arithmetic mean of Specificity (True Negative Rate) and Sensitivity (True Positive Rate). ii) Recall: measures the proportion of actual positive cases that the model correctly identified. iii) Cumulative Statistical Parity Difference (CumSPD): It is a measure of the cumulative statistical parity difference between the protected and deprived groups (Equation \ref{equ:Statistical Parity Difference}). This measure must be equal to 0 to be fair. iv) Cumulative Equal Opportunity Difference (CumEOD): It is a measure of cumulative equal opportunity difference between the privileged and unprivileged groups (Equation \ref{equ:Equal_Opportunity_Difference}).

 \subsection{Experimental Results}
 \label{sec:Result}

 \subsubsection{$FS^2$ compared to state-of-the-art methods}

We compare the performance-fairness metric of our proposed method ($FS^2$) against the four baselines. The results are demonstrated in Table ~\ref{tab:result}, the darker hue of red denotes the best model performance followed by a lighter red/ pink shade that signifies the second-best performance. 
$FS^2$ performed the best in 12 of the 16 performance-fairness evaluations. It was a close second (with a margin of 1-3\%) in the remaining 4 measurements. 
\begin{wrapfigure}{r}{0.4\textwidth}
	\centering
	\includegraphics[width=0.4\textwidth]{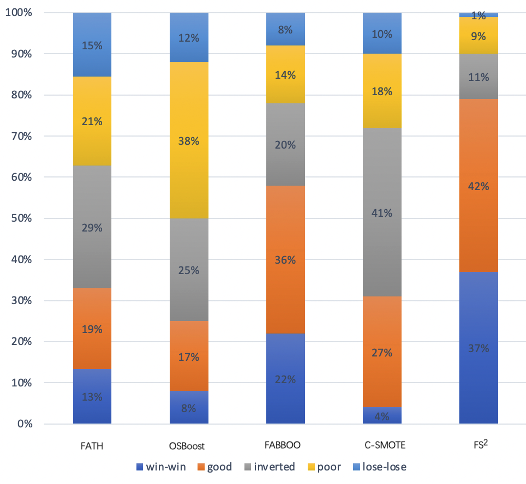}\vspace{-0.3cm}
	\caption{$FS^2$ and other methods' effectiveness distribution in benchmark tasks; $FS^2$ shows the best balance, with 79\% of mitigation cases having good or win-win results. \vspace{-0.3cm}}
	\label{fig:RQ1}
\end{wrapfigure}
In the BNK dataset, $FS^2$ achieved a maximum Balanced Accuracy of 81\%, surpassing the powerful C-SMOTE. The cumulative EOD and SPD were the best for $FS^2$ with a 0.02 and 0.01 score respectively, while its recall score of 0.78 was the second highest.
$FS^2$ outperformed these models on the GMSC dataset in terms of Balanced Accuracy, Cumulative EOD and SPD with respective scores of 0.83, 0.01 and 0.01. It was second in terms of Recall with a score of 0.78.
Similarly, for the NYPD dataset $FS^2$ surpassed all four baseline methods as it got the best score for all the fairness-performance metrics i.e. Balanced Accuracy (0.68), Recall (0.54),  Cumulative SPD (0.06) and Cumulative EOD (0.03).
For the SYN dataset, $FS^2$ performed fairly well--it was the second best in terms of Balanced Accuracy and Recall. But it outshined in terms of the fairness metrics as it scored a cumulative SPD (0.04) and OPD (0.06). 

To summarize, $FS^2$ demonstrates superior performance compared to other methods in terms of both accuracy and fairness. 



\subsubsection{The Trade-Off of Effectiveness between $FS^2$ and other State-of-the-Art Methods}

 With the proposed FBU, we can now evaluate the trade-off between fairness and performance based on one single illustrative metric. Figure \ref{fig:RQ1} shows the overall results. As we can see, $FS^2$ achieves a good or win-win trade-off (i.e., it beats the trade-off baseline constructed by FBU) in most cases, i.e., 79\% of the time. 

In comparison, the corresponding percentages for FATH, OSBoost, FABBOO, and C-SMOTE were 32\%, 25\%, 58\%, 31\%, respectively. In addition, $FS^2$ has significantly fewer lose-lose trade-off cases (just 1\%) than other existing methods. For example, FATH has a lose-lose trade-off rate of 15\%, 15 times higher than that of $FS^2$. Overall, $FS^2$ achieves the best trade-off, and it is a significant improvement over all existing methods. The obtained results in Figure \ref{fig:RQ1} are also consistent with the results in Table~\ref{tab:result}, but are presented in a unified manner that is easy to interpret, verifying the theoretical design of FBU.

\begin{table}
    \centering
        \caption{Overall predictive and fairness performance for the (darker cells show top rank and the lighter cells show the second rank).}
    \begin{tabular}{|c|c|c|c|c|c|c|}
    \hline
    Dataset&\makecell[c]{Sensitive \\ Attribute }&Methods &\makecell[c]{Balanced \\ Accuracy }&	Recall & \makecell[c]{Cumulative \\ SPD}&	\makecell[c]{Cumulative \\ EOD}	\\
    \hline 
    \multirow{5}{*}{BNK}&\multirow{5}{*}{Age}&FAHT & 0.73&	0.52&0.16&	0.18\\
    \cline{3-7} 
&&OSBoost&	0.73&	0.71&		0.18&	0.19\\
\cline{3-7}
&&FABBOO&	0.76&	0.54&		\cellcolor{red!15}0.05&	\cellcolor{red!15}0.08\\
\cline{3-7}
&&C-SMOTE&	\cellcolor{red!15}0.80&	\cellcolor{red!40}0.81&		0.34&	0.14\\
\cline{3-7}
&&$FS^2$&	\cellcolor{red!40}0.81 &	\cellcolor{red!15}0.78&		\cellcolor{red!40}0.02 &	\cellcolor{red!40}0.01\\
\hline

   \multirow{5}{*}{GMSC}&\multirow{5}{*}{Age}&FAHT & 0.62&	0.28&\cellcolor{red!15}0.02&	0.04\\
    \cline{3-7} 
&&OSBoost&	0.65&	0.57&		0.04&	0.06\\
\cline{3-7}
&&FABBOO&	0.81&	0.76&		\cellcolor{red!15}0.02&	\cellcolor{red!40}0.01\\
\cline{3-7}
&&C-SMOTE&	\cellcolor{red!15}0.80&	\cellcolor{red!40}0.80&		0.07&	0.02\\
\cline{3-7}
&&$FS^2$&	\cellcolor{red!40}0.83 &	\cellcolor{red!15}0.79&		\cellcolor{red!40}0.01 &	\cellcolor{red!40}0.01\\
\hline

   \multirow{5}{*}{NYPD}&\multirow{5}{*}{Gender}&FAHT & 0.55&	0.28&0.17&	0.28\\
    \cline{3-7} 
&&OSBoost&	0.52&	0.07&		0.25&	0.15\\
\cline{3-7}
&&FABBOO&	\cellcolor{red!15}0.62&	\cellcolor{red!15}0.50&		\cellcolor{red!15}0.07&	\cellcolor{red!15}0.06\\
\cline{3-7}
&&C-SMOTE&	0.56&	0.42&		0.34&	0.16\\
\cline{3-7}
&&$FS^2$&	\cellcolor{red!40}0.68 &	\cellcolor{red!40}0.54&		\cellcolor{red!40}0.06 &	\cellcolor{red!40}0.03\\
\hline

   \multirow{5}{*}{SYN}&\multirow{5}{*}{synth.}&FAHT & 0.62&	0.28&0.08&	0.16\\
    \cline{3-7} 
&&OSBoost&	0.57&	0.31&		0.08&	0.11\\
\cline{3-7}
&&FABBOO&	\cellcolor{red!40}0.67&	0.58&		\cellcolor{red!15}0.09&	\cellcolor{red!15}0.07\\
\cline{3-7}
&&C-SMOTE&	\cellcolor{red!40}0.67&	\cellcolor{red!40}0.62&		0.26&	0.16\\
\cline{3-7}
&&$FS^2$&	\cellcolor{red!15}0.65 &	\cellcolor{red!15}0.61&		\cellcolor{red!40}0.04 &	\cellcolor{red!40}0.05\\
\hline

    \end{tabular}

    \label{tab:result}
\end{table}

\subsubsection{Influence of different decay factors}

In this section, we analyze the effect of varying decay factor settings on model fairness and performance in the presence of diverse data flow variations. i) Fixed Class-imbalance: In this situation the class ratio is fixed over the data stream. We noted that low decay factor values negatively impact model performance, with Recall dropping below 60\% for decay factor $\textless$ 0.4. However, for decay factor $\geq$ 0.5, Balanced Accuracy and Recall were not significantly impacted. Furthermore, the model's capacity to alleviate unfair outcomes was unchanged for all decay factor values. ii) Increasing and decreasing class imbalance: we observed a steady rise in Recall as the decay factor increased. iii) Fluctuating Class-imbalance: low values of decay factor decrease the performance as in previous cases, while high values also hurt the minority class. In this case, the positive class alternates between minority and majority. High decay factor values allow the model to consider a greater amount of history, leading to class imbalance weights assigned by $FS^2$ not being adjusted to recent changes. Overall, the value of the decay factor directly affects the performance of $FS^2$. Low values of the decay factor produce fluctuating class imbalance weights that decrease the model's performance, and in some cases, values close to 1 are also not appropriate.

\begin{figure}[!htbp]
	\centering
	\includegraphics[width=1\textwidth, height=0.25\textwidth]{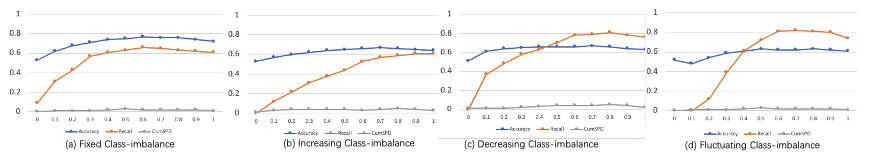}
	\caption{Impact of decay factor on the synthetic dataset of varying class imbalance}
	\label{fig:RQ3-1}
\end{figure}

\subsubsection{Is $FS^2$ capable of handling concept drift?}
\begin{wrapfigure}{r}{0.4\textwidth}
	\centering
	\includegraphics[width=0.4\textwidth]{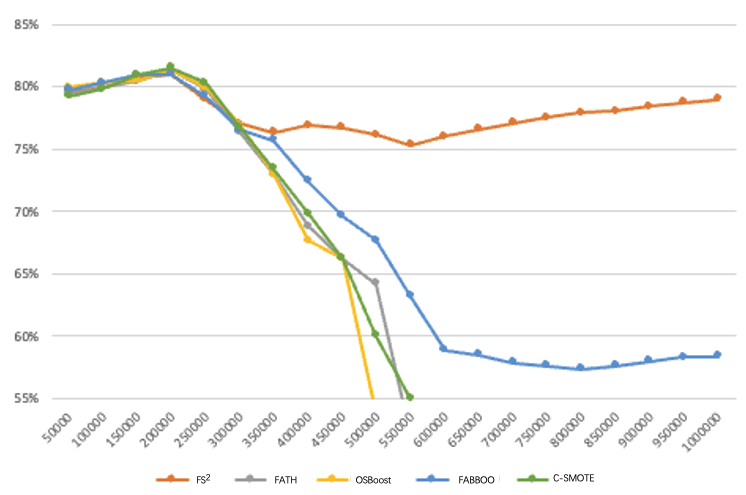}\vspace{-0.3cm}
	\caption{Proportion of mitigation cases that beat the trade-off baseline of $FS^2$ and existing methods in synthesis dataset with concept drift. \vspace{-0.3cm}}
	\label{fig:RQ4}
\end{wrapfigure}

We tracked the variation of the model performance on the simulated dataset using FBU, and the results are shown in Figure~\ref{fig:RQ4}. Our approach (the orange curve, towards the top of the graph) is designed for streaming data, and we do observe that the model has the capability to adapt and regain its performance after a concept drift takes place (the curve goes down and then goes up). And since we can handle fairness drifts, we observe that our model can adapt to both fairness drift and conceptual drift. For example, C-SMOTE has a significant decrease in the overall performance of the model because of the lack of detection of fairness drift. At the same time, the intuition and validity of FBU are well represented. Before BFU, we would need to show the changes in performance and fairness separately, and it would be difficult to analyze the specific impact tradeoff between them. FBU makes this analysis far simpler. 

\subsubsection{Effect of varying Window size}

\begin{wrapfigure}{r}{0.4\textwidth}
	\centering
	\includegraphics[width=0.4\textwidth]{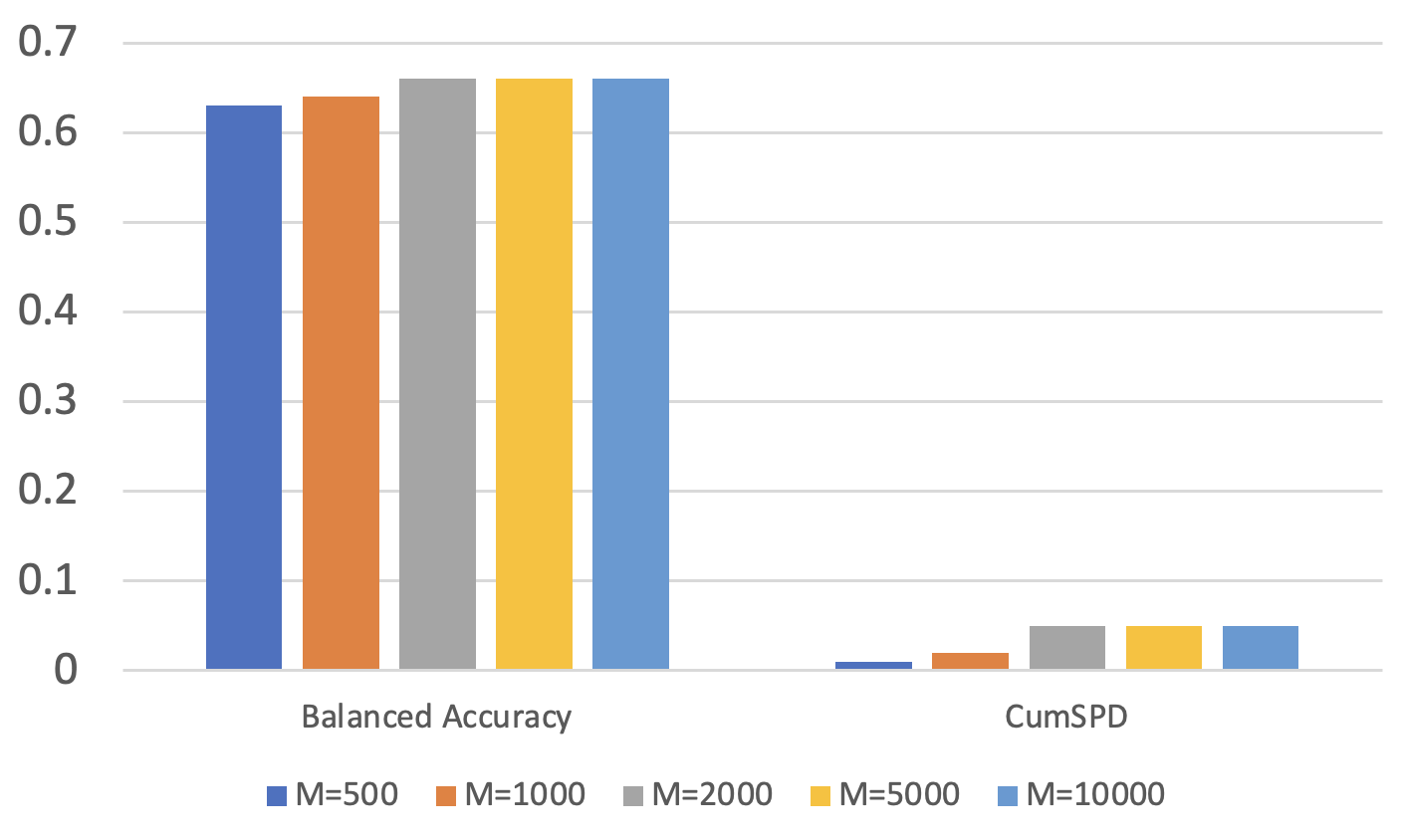}\vspace{-0.3cm}
	\caption{Effect of window size on model performance and fairness. \vspace{-0.3cm}}
	\label{fig:RQ5}
\end{wrapfigure}

In our study, we evaluated the impact of various window sizes on our method, specifically using window sizes of 500, 1000, 2000, 5000, and 10000. For predictive performance, we present results on Balanced Accuracy, and for fairness we report CumSPD. As Figure~\ref{fig:RQ5} shows, we observe that when the window size surpasses 2000, Balanced Accuracy and CumSPD remain constant. This is due to the occurrence of concept drift, the large sliding window (window size > 2000) is not able to discard outdated instances from before the concept drift, while the smaller window (size = 500 or 1000) is able to accommodate the newer instances after concept drift and discards the older ones. Generally, smaller window settings yield more fair results, but at the cost of decreased prediction performance. Conversely, excessively large window settings can impede the model's ability to adjust decision boundaries during concept drift, thus negatively affecting model fairness.

\section{Conclusion}
\label{sec:conclusion}

Ensuring models make fair predictions in the presence of a continuous infinite stream of data is an open challenge in real-world streaming machine learning applications. In this work, we proposed Fair Bonded Utility ($FBU$), the first fairness metric to unify the comparison of the fairness-performance trade-offs of multiple fairness techniques into one intuitive evaluation that simplifies the task of choosing a fairness technique. Then we presented Fair Sampling over Stream ($FC^2$), a novel fair class-balancing technique capable of handling continuous data streams with concept drift. Extensive experiments show our approach gives superior fairness results according to multiple widely-used fairness metrics without sacrificing performance. The pre-processing nature of our method can help improve performance and fairness in many varied applications due to ease of integration.

\balance
\bibliographystyle{ACM-Reference-Format}
\bibliography{typeinst}


\begin{thebibliography}{58}


\ifx \showCODEN    \undefined \def \showCODEN     #1{\unskip}     \fi
\ifx \showDOI      \undefined \def \showDOI       #1{#1}\fi
\ifx \showISBNx    \undefined \def \showISBNx     #1{\unskip}     \fi
\ifx \showISBNxiii \undefined \def \showISBNxiii  #1{\unskip}     \fi
\ifx \showISSN     \undefined \def \showISSN      #1{\unskip}     \fi
\ifx \showLCCN     \undefined \def \showLCCN      #1{\unskip}     \fi
\ifx \shownote     \undefined \def \shownote      #1{#1}          \fi
\ifx \showarticletitle \undefined \def \showarticletitle #1{#1}   \fi
\ifx \showURL      \undefined \def \showURL       {\relax}        \fi
\providecommand\bibfield[2]{#2}
\providecommand\bibinfo[2]{#2}
\providecommand\natexlab[1]{#1}
\providecommand\showeprint[2][]{arXiv:#2}

\bibitem[Aggarwal(2007)]%
        {aggarwal2007data}
\bibfield{author}{\bibinfo{person}{Charu~C Aggarwal}.}
  \bibinfo{year}{2007}\natexlab{}.
\newblock \bibinfo{booktitle}{\emph{Data streams: models and algorithms}}.
  Vol.~\bibinfo{volume}{31}.
\newblock \bibinfo{publisher}{Springer Science \& Business Media}.
\newblock


\bibitem[Bechavod et~al\mbox{.}(2020)]%
        {bechavod2020metric}
\bibfield{author}{\bibinfo{person}{Yahav Bechavod},
  \bibinfo{person}{Christopher Jung}, {and} \bibinfo{person}{Steven~Z Wu}.}
  \bibinfo{year}{2020}\natexlab{}.
\newblock \showarticletitle{Metric-free individual fairness in online
  learning}.
\newblock \bibinfo{journal}{\emph{Advances in neural information processing
  systems}}  \bibinfo{volume}{33} (\bibinfo{year}{2020}),
  \bibinfo{pages}{11214--11225}.
\newblock


\bibitem[Bernardo et~al\mbox{.}(2020)]%
        {bernardo2020c}
\bibfield{author}{\bibinfo{person}{Alessio Bernardo},
  \bibinfo{person}{Heitor~Murilo Gomes}, \bibinfo{person}{Jacob Montiel},
  \bibinfo{person}{Bernhard Pfahringer}, \bibinfo{person}{Albert Bifet}, {and}
  \bibinfo{person}{Emanuele Della~Valle}.} \bibinfo{year}{2020}\natexlab{}.
\newblock \showarticletitle{C-smote: Continuous synthetic minority oversampling
  for evolving data streams}. In \bibinfo{booktitle}{\emph{2020 IEEE
  International Conference on Big Data (Big Data)}}. IEEE,
  \bibinfo{pages}{483--492}.
\newblock


\bibitem[Bifet and Gavalda(2009)]%
        {bifet2009adaptive}
\bibfield{author}{\bibinfo{person}{Albert Bifet} {and} \bibinfo{person}{Ricard
  Gavalda}.} \bibinfo{year}{2009}\natexlab{}.
\newblock \showarticletitle{Adaptive learning from evolving data streams}. In
  \bibinfo{booktitle}{\emph{Advances in Intelligent Data Analysis VIII: 8th
  International Symposium on Intelligent Data Analysis, IDA 2009, Lyon, France,
  August 31-September 2, 2009. Proceedings 8}}. Springer,
  \bibinfo{pages}{249--260}.
\newblock


\bibitem[Bifet and Gavaldà(2007)]%
        {inproceedings}
\bibfield{author}{\bibinfo{person}{Albert Bifet} {and} \bibinfo{person}{Ricard
  Gavaldà}.} \bibinfo{year}{2007}\natexlab{}.
\newblock \showarticletitle{Learning from Time-Changing Data with Adaptive
  Windowing}.
\newblock \bibinfo{journal}{\emph{Proceedings of the 7th SIAM International
  Conference on Data Mining}}  \bibinfo{volume}{7}.
\newblock
\urldef\tempurl%
\url{https://doi.org/10.1137/1.9781611972771.42}
\showDOI{\tempurl}


\bibitem[Calders et~al\mbox{.}(2009)]%
        {calders2009building}
\bibfield{author}{\bibinfo{person}{Toon Calders}, \bibinfo{person}{Faisal
  Kamiran}, {and} \bibinfo{person}{Mykola Pechenizkiy}.}
  \bibinfo{year}{2009}\natexlab{}.
\newblock \showarticletitle{Building classifiers with independency
  constraints}. In \bibinfo{booktitle}{\emph{2009 IEEE international conference
  on data mining workshops}}. IEEE, \bibinfo{pages}{13--18}.
\newblock


\bibitem[Calders and Verwer(2010)]%
        {calders2010three}
\bibfield{author}{\bibinfo{person}{Toon Calders} {and} \bibinfo{person}{Sicco
  Verwer}.} \bibinfo{year}{2010}\natexlab{}.
\newblock \showarticletitle{Three naive bayes approaches for
  discrimination-free classification}.
\newblock \bibinfo{journal}{\emph{Data mining and knowledge discovery}}
  \bibinfo{volume}{21} (\bibinfo{year}{2010}), \bibinfo{pages}{277--292}.
\newblock


\bibitem[Chakraborty et~al\mbox{.}(2021)]%
        {chakraborty2021bias}
\bibfield{author}{\bibinfo{person}{Joymallya Chakraborty},
  \bibinfo{person}{Suvodeep Majumder}, {and} \bibinfo{person}{Tim Menzies}.}
  \bibinfo{year}{2021}\natexlab{}.
\newblock \showarticletitle{Bias in machine learning software: Why? how? what
  to do?}. In \bibinfo{booktitle}{\emph{Proceedings of the 29th ACM Joint
  Meeting on European Software Engineering Conference and Symposium on the
  Foundations of Software Engineering}}. \bibinfo{pages}{429--440}.
\newblock


\bibitem[Chawla et~al\mbox{.}(2002)]%
        {chawla2002smote}
\bibfield{author}{\bibinfo{person}{Nitesh~V Chawla}, \bibinfo{person}{Kevin~W
  Bowyer}, \bibinfo{person}{Lawrence~O Hall}, {and} \bibinfo{person}{W~Philip
  Kegelmeyer}.} \bibinfo{year}{2002}\natexlab{}.
\newblock \showarticletitle{SMOTE: synthetic minority over-sampling technique}.
\newblock \bibinfo{journal}{\emph{Journal of artificial intelligence research}}
   \bibinfo{volume}{16} (\bibinfo{year}{2002}), \bibinfo{pages}{321--357}.
\newblock


\bibitem[Chen et~al\mbox{.}(2012)]%
        {chen2012online}
\bibfield{author}{\bibinfo{person}{Shang-Tse Chen}, \bibinfo{person}{Hsuan-Tien
  Lin}, {and} \bibinfo{person}{Chi-Jen Lu}.} \bibinfo{year}{2012}\natexlab{}.
\newblock \showarticletitle{An online boosting algorithm with theoretical
  justifications}.
\newblock \bibinfo{journal}{\emph{arXiv preprint arXiv:1206.6422}}
  (\bibinfo{year}{2012}).
\newblock


\bibitem[Crawford(2016)]%
        {crawford2016artificial}
\bibfield{author}{\bibinfo{person}{Kate Crawford}.}
  \bibinfo{year}{2016}\natexlab{}.
\newblock \showarticletitle{Artificial intelligence’s white guy problem}.
\newblock \bibinfo{journal}{\emph{The New York Times}} \bibinfo{volume}{25},
  \bibinfo{number}{06} (\bibinfo{year}{2016}), \bibinfo{pages}{5}.
\newblock


\bibitem[Datta et~al\mbox{.}(2014)]%
        {datta2014automated}
\bibfield{author}{\bibinfo{person}{Amit Datta}, \bibinfo{person}{Michael~Carl
  Tschantz}, {and} \bibinfo{person}{Anupam Datta}.}
  \bibinfo{year}{2014}\natexlab{}.
\newblock \showarticletitle{Automated experiments on ad privacy settings: A
  tale of opacity, choice, and discrimination}.
\newblock \bibinfo{journal}{\emph{arXiv preprint arXiv:1408.6491}}
  (\bibinfo{year}{2014}).
\newblock


\bibitem[Douzas et~al\mbox{.}(2018)]%
        {douzas2018improving}
\bibfield{author}{\bibinfo{person}{Georgios Douzas}, \bibinfo{person}{Fernando
  Bacao}, {and} \bibinfo{person}{Felix Last}.} \bibinfo{year}{2018}\natexlab{}.
\newblock \showarticletitle{Improving imbalanced learning through a heuristic
  oversampling method based on k-means and SMOTE}.
\newblock \bibinfo{journal}{\emph{Information Sciences}}  \bibinfo{volume}{465}
  (\bibinfo{year}{2018}), \bibinfo{pages}{1--20}.
\newblock


\bibitem[Gama(2010)]%
        {gama2010knowledge}
\bibfield{author}{\bibinfo{person}{Joao Gama}.}
  \bibinfo{year}{2010}\natexlab{}.
\newblock \bibinfo{booktitle}{\emph{Knowledge discovery from data streams}}.
\newblock \bibinfo{publisher}{CRC Press}.
\newblock


\bibitem[Gama et~al\mbox{.}(2014)]%
        {gama2014survey}
\bibfield{author}{\bibinfo{person}{Jo{\~a}o Gama}, \bibinfo{person}{Indr{\.e}
  {\v{Z}}liobait{\.e}}, \bibinfo{person}{Albert Bifet}, \bibinfo{person}{Mykola
  Pechenizkiy}, {and} \bibinfo{person}{Abdelhamid Bouchachia}.}
  \bibinfo{year}{2014}\natexlab{}.
\newblock \showarticletitle{A survey on concept drift adaptation}.
\newblock \bibinfo{journal}{\emph{ACM computing surveys (CSUR)}}
  \bibinfo{volume}{46}, \bibinfo{number}{4} (\bibinfo{year}{2014}),
  \bibinfo{pages}{44}.
\newblock


\bibitem[Ghazikhani et~al\mbox{.}(2014)]%
        {ghazikhani2014online}
\bibfield{author}{\bibinfo{person}{Adel Ghazikhani}, \bibinfo{person}{Reza
  Monsefi}, {and} \bibinfo{person}{Hadi Sadoghi~Yazdi}.}
  \bibinfo{year}{2014}\natexlab{}.
\newblock \showarticletitle{Online neural network model for non-stationary and
  imbalanced data stream classification}.
\newblock \bibinfo{journal}{\emph{International Journal of Machine Learning and
  Cybernetics}}  \bibinfo{volume}{5} (\bibinfo{year}{2014}),
  \bibinfo{pages}{51--62}.
\newblock


\bibitem[Ghazikhani et~al\mbox{.}(2013)]%
        {ghazikhani2013ensemble}
\bibfield{author}{\bibinfo{person}{Adel Ghazikhani}, \bibinfo{person}{Reza
  Monsefi}, {and} \bibinfo{person}{Hadi~Sadoghi Yazdi}.}
  \bibinfo{year}{2013}\natexlab{}.
\newblock \showarticletitle{Ensemble of online neural networks for
  non-stationary and imbalanced data streams}.
\newblock \bibinfo{journal}{\emph{Neurocomputing}}  \bibinfo{volume}{122}
  (\bibinfo{year}{2013}), \bibinfo{pages}{535--544}.
\newblock


\bibitem[Gomes et~al\mbox{.}(2017)]%
        {gomes2017adaptive}
\bibfield{author}{\bibinfo{person}{Heitor~M Gomes}, \bibinfo{person}{Albert
  Bifet}, \bibinfo{person}{Jesse Read}, \bibinfo{person}{Jean~Paul Barddal},
  \bibinfo{person}{Fabr{\'\i}cio Enembreck}, \bibinfo{person}{Bernhard
  Pfharinger}, \bibinfo{person}{Geoff Holmes}, {and} \bibinfo{person}{Talel
  Abdessalem}.} \bibinfo{year}{2017}\natexlab{}.
\newblock \showarticletitle{Adaptive random forests for evolving data stream
  classification}.
\newblock \bibinfo{journal}{\emph{Machine Learning}}  \bibinfo{volume}{106}
  (\bibinfo{year}{2017}), \bibinfo{pages}{1469--1495}.
\newblock


\bibitem[Hajian et~al\mbox{.}(2016)]%
        {hajian2016algorithmic}
\bibfield{author}{\bibinfo{person}{Sara Hajian}, \bibinfo{person}{Francesco
  Bonchi}, {and} \bibinfo{person}{Carlos Castillo}.}
  \bibinfo{year}{2016}\natexlab{}.
\newblock \showarticletitle{Algorithmic bias: From discrimination discovery to
  fairness-aware data mining}. In \bibinfo{booktitle}{\emph{Proceedings of the
  22nd ACM SIGKDD international conference on knowledge discovery and data
  mining}}. \bibinfo{pages}{2125--2126}.
\newblock


\bibitem[Hardt et~al\mbox{.}(2016)]%
        {hardt2016equality}
\bibfield{author}{\bibinfo{person}{Moritz Hardt}, \bibinfo{person}{Eric Price},
  {and} \bibinfo{person}{Nati Srebro}.} \bibinfo{year}{2016}\natexlab{}.
\newblock \showarticletitle{Equality of opportunity in supervised learning}.
\newblock \bibinfo{journal}{\emph{Advances in neural information processing
  systems}}  \bibinfo{volume}{29} (\bibinfo{year}{2016}).
\newblock


\bibitem[https://www.kaggle.com/c/GiveMeSomeCredit/overview(2021)]%
        {https://www.kaggle.com/c/givemesomecredit/overview_2021}
\bibfield{author}{\bibinfo{person}{https://www.kaggle.com/c/GiveMeSomeCredit/overview}.}
  \bibinfo{year}{2021}\natexlab{}.
\newblock \bibinfo{title}{Give Me Some Credit}.
\newblock
\newblock
\urldef\tempurl%
\url{https://doi.org/10.34740/KAGGLE/DSV/2242482}
\showDOI{\tempurl}


\bibitem[Iosifidis et~al\mbox{.}(2019)]%
        {iosifidis2019fairness}
\bibfield{author}{\bibinfo{person}{Vasileios Iosifidis}, \bibinfo{person}{Thi
  Ngoc~Han Tran}, {and} \bibinfo{person}{Eirini Ntoutsi}.}
  \bibinfo{year}{2019}\natexlab{}.
\newblock \showarticletitle{Fairness-enhancing interventions in stream
  classification}. In \bibinfo{booktitle}{\emph{Database and Expert Systems
  Applications: 30th International Conference, DEXA 2019, Linz, Austria, August
  26--29, 2019, Proceedings, Part I 30}}. Springer, \bibinfo{pages}{261--276}.
\newblock


\bibitem[Iosifidis et~al\mbox{.}(2021)]%
        {iosifidis2021online}
\bibfield{author}{\bibinfo{person}{Vasileios Iosifidis},
  \bibinfo{person}{Wenbin Zhang}, {and} \bibinfo{person}{Eirini Ntoutsi}.}
  \bibinfo{year}{2021}\natexlab{}.
\newblock \showarticletitle{Online fairness-aware learning with imbalanced data
  streams}.
\newblock \bibinfo{journal}{\emph{arXiv preprint arXiv:2108.06231}}
  (\bibinfo{year}{2021}).
\newblock


\bibitem[Kamiran and Calders(2009)]%
        {kamiran2009classifying}
\bibfield{author}{\bibinfo{person}{Faisal Kamiran} {and} \bibinfo{person}{Toon
  Calders}.} \bibinfo{year}{2009}\natexlab{}.
\newblock \showarticletitle{Classifying without discriminating}. In
  \bibinfo{booktitle}{\emph{2009 2nd international conference on computer,
  control and communication}}. IEEE, \bibinfo{pages}{1--6}.
\newblock


\bibitem[Kamiran and Calders(2012)]%
        {kamiran2012data}
\bibfield{author}{\bibinfo{person}{Faisal Kamiran} {and} \bibinfo{person}{Toon
  Calders}.} \bibinfo{year}{2012}\natexlab{}.
\newblock \showarticletitle{Data preprocessing techniques for classification
  without discrimination}.
\newblock \bibinfo{journal}{\emph{Knowledge and information systems}}
  \bibinfo{volume}{33}, \bibinfo{number}{1} (\bibinfo{year}{2012}),
  \bibinfo{pages}{1--33}.
\newblock


\bibitem[Kamiran et~al\mbox{.}(2010)]%
        {kamiran2010discrimination}
\bibfield{author}{\bibinfo{person}{Faisal Kamiran}, \bibinfo{person}{Toon
  Calders}, {and} \bibinfo{person}{Mykola Pechenizkiy}.}
  \bibinfo{year}{2010}\natexlab{}.
\newblock \showarticletitle{Discrimination aware decision tree learning}. In
  \bibinfo{booktitle}{\emph{2010 IEEE international conference on data
  mining}}. IEEE, \bibinfo{pages}{869--874}.
\newblock


\bibitem[Krawczyk et~al\mbox{.}(2017)]%
        {krawczyk2017ensemble}
\bibfield{author}{\bibinfo{person}{Bartosz Krawczyk},
  \bibinfo{person}{Leandro~L Minku}, \bibinfo{person}{Jo{\~a}o Gama},
  \bibinfo{person}{Jerzy Stefanowski}, {and} \bibinfo{person}{Micha{\l}
  Wo{\'z}niak}.} \bibinfo{year}{2017}\natexlab{}.
\newblock \showarticletitle{Ensemble learning for data stream analysis: A
  survey}.
\newblock \bibinfo{journal}{\emph{Information Fusion}}  \bibinfo{volume}{37}
  (\bibinfo{year}{2017}), \bibinfo{pages}{132--156}.
\newblock


\bibitem[Lee et~al\mbox{.}(2019)]%
        {lee2019algorithmic}
\bibfield{author}{\bibinfo{person}{Nicol~Turner Lee}, \bibinfo{person}{Paul
  Resnick}, {and} \bibinfo{person}{Genie Barton}.}
  \bibinfo{year}{2019}\natexlab{}.
\newblock \showarticletitle{Algorithmic bias detection and mitigation: Best
  practices and policies to reduce consumer harms}.
\newblock \bibinfo{journal}{\emph{Brookings Institute: Washington, DC, USA}}
  \bibinfo{volume}{2} (\bibinfo{year}{2019}).
\newblock


\bibitem[Moro et~al\mbox{.}(2014)]%
        {MORO201422}
\bibfield{author}{\bibinfo{person}{Sérgio Moro}, \bibinfo{person}{Paulo
  Cortez}, {and} \bibinfo{person}{Paulo Rita}.}
  \bibinfo{year}{2014}\natexlab{}.
\newblock \showarticletitle{A data-driven approach to predict the success of
  bank telemarketing}.
\newblock \bibinfo{journal}{\emph{Decision Support Systems}}
  \bibinfo{volume}{62} (\bibinfo{year}{2014}), \bibinfo{pages}{22--31}.
\newblock
\showISSN{0167-9236}
\urldef\tempurl%
\url{https://doi.org/10.1016/j.dss.2014.03.001}
\showDOI{\tempurl}


\bibitem[Prates et~al\mbox{.}(2020)]%
        {prates2020assessing}
\bibfield{author}{\bibinfo{person}{Marcelo~OR Prates}, \bibinfo{person}{Pedro~H
  Avelar}, {and} \bibinfo{person}{Lu{\'\i}s~C Lamb}.}
  \bibinfo{year}{2020}\natexlab{}.
\newblock \showarticletitle{Assessing gender bias in machine translation: a
  case study with google translate}.
\newblock \bibinfo{journal}{\emph{Neural Computing and Applications}}
  \bibinfo{volume}{32} (\bibinfo{year}{2020}), \bibinfo{pages}{6363--6381}.
\newblock


\bibitem[Quy et~al\mbox{.}(2022)]%
        {li2021time}
\bibfield{author}{\bibinfo{person}{Tai~Le Quy}, \bibinfo{person}{Arjun Roy},
  \bibinfo{person}{Vasileios Iosifidis}, \bibinfo{person}{Wenbin Zhang}, {and}
  \bibinfo{person}{Eirini Ntoutsi}.} \bibinfo{year}{2022}\natexlab{}.
\newblock \showarticletitle{A survey on datasets for fairness-aware machine
  learning}.
\newblock \bibinfo{journal}{\emph{Data Mining and Knowledge Discovery}}
  (\bibinfo{year}{2022}).
\newblock


\bibitem[Read et~al\mbox{.}(2019)]%
        {read2019error}
\bibfield{author}{\bibinfo{person}{Jesse Read}, \bibinfo{person}{Nikolaos
  Tziortziotis}, {and} \bibinfo{person}{Michalis Vazirgiannis}.}
  \bibinfo{year}{2019}\natexlab{}.
\newblock \showarticletitle{Error-space representations for multi-dimensional
  data streams with temporal dependence}.
\newblock \bibinfo{journal}{\emph{Pattern Analysis and Applications}}
  \bibinfo{volume}{22}, \bibinfo{number}{3} (\bibinfo{year}{2019}),
  \bibinfo{pages}{1211--1220}.
\newblock


\bibitem[Sargent and Shea(2017)]%
        {sargent2017office}
\bibfield{author}{\bibinfo{person}{John~F Sargent} {and}
  \bibinfo{person}{Dana~A Shea}.} \bibinfo{year}{2017}\natexlab{}.
\newblock \bibinfo{booktitle}{\emph{Office of Science and Technology Policy
  (OSTP): History and Overview}}.
\newblock \bibinfo{publisher}{Congressional Research Service}.
\newblock


\bibitem[Sharma(2017)]%
        {sharma2017pros}
\bibfield{author}{\bibinfo{person}{Gaganpreet Sharma}.}
  \bibinfo{year}{2017}\natexlab{}.
\newblock \showarticletitle{Pros and cons of different sampling techniques}.
\newblock \bibinfo{journal}{\emph{International journal of applied research}}
  \bibinfo{volume}{3}, \bibinfo{number}{7} (\bibinfo{year}{2017}),
  \bibinfo{pages}{749--752}.
\newblock


\bibitem[Sun et~al\mbox{.}(2020)]%
        {sun2020automatic}
\bibfield{author}{\bibinfo{person}{Zeyu Sun}, \bibinfo{person}{Jie~M Zhang},
  \bibinfo{person}{Mark Harman}, \bibinfo{person}{Mike Papadakis}, {and}
  \bibinfo{person}{Lu Zhang}.} \bibinfo{year}{2020}\natexlab{}.
\newblock \showarticletitle{Automatic testing and improvement of machine
  translation}. In \bibinfo{booktitle}{\emph{Proceedings of the ACM/IEEE 42nd
  International Conference on Software Engineering}}.
  \bibinfo{pages}{974--985}.
\newblock


\bibitem[Sweeney(2013)]%
        {sweeney2013discrimination}
\bibfield{author}{\bibinfo{person}{Latanya Sweeney}.}
  \bibinfo{year}{2013}\natexlab{}.
\newblock \showarticletitle{Discrimination in online ad delivery}.
\newblock \bibinfo{journal}{\emph{Commun. ACM}} \bibinfo{volume}{56},
  \bibinfo{number}{5} (\bibinfo{year}{2013}), \bibinfo{pages}{44--54}.
\newblock


\bibitem[Tang et~al\mbox{.}(2020)]%
        {tang2020using}
\bibfield{author}{\bibinfo{person}{Xuejiao Tang}, \bibinfo{person}{Liuhua
  Zhang}, {et~al\mbox{.}}} \bibinfo{year}{2020}\natexlab{}.
\newblock \showarticletitle{Using machine learning to automate mammogram images
  analysis}. In \bibinfo{booktitle}{\emph{IEEE International Conference on
  Bioinformatics and Biomedicine (BIBM)}}. \bibinfo{pages}{757--764}.
\newblock


\bibitem[Verma and Rubin(2018)]%
        {verma2018fairness}
\bibfield{author}{\bibinfo{person}{Sahil Verma} {and} \bibinfo{person}{Julia
  Rubin}.} \bibinfo{year}{2018}\natexlab{}.
\newblock \showarticletitle{Fairness definitions explained}. In
  \bibinfo{booktitle}{\emph{Proceedings of the international workshop on
  software fairness}}. \bibinfo{pages}{1--7}.
\newblock


\bibitem[Wang and Pineau(2016)]%
        {wang2016online}
\bibfield{author}{\bibinfo{person}{Boyu Wang} {and} \bibinfo{person}{Joelle
  Pineau}.} \bibinfo{year}{2016}\natexlab{}.
\newblock \showarticletitle{Online bagging and boosting for imbalanced data
  streams}.
\newblock \bibinfo{journal}{\emph{IEEE Transactions on Knowledge and Data
  Engineering}} \bibinfo{volume}{28}, \bibinfo{number}{12}
  (\bibinfo{year}{2016}), \bibinfo{pages}{3353--3366}.
\newblock


\bibitem[Wang et~al\mbox{.}(2020)]%
        {wang2020visual}
\bibfield{author}{\bibinfo{person}{Qianwen Wang}, \bibinfo{person}{Zhenhua Xu},
  \bibinfo{person}{Zhutian Chen}, \bibinfo{person}{Yong Wang},
  \bibinfo{person}{Shixia Liu}, {and} \bibinfo{person}{Huamin Qu}.}
  \bibinfo{year}{2020}\natexlab{}.
\newblock \showarticletitle{Visual analysis of discrimination in machine
  learning}.
\newblock \bibinfo{journal}{\emph{IEEE Transactions on Visualization and
  Computer Graphics}} \bibinfo{volume}{27}, \bibinfo{number}{2}
  (\bibinfo{year}{2020}), \bibinfo{pages}{1470--1480}.
\newblock


\bibitem[Wang et~al\mbox{.}(2013)]%
        {wang2013learning}
\bibfield{author}{\bibinfo{person}{Shuo Wang}, \bibinfo{person}{Leandro~L
  Minku}, {and} \bibinfo{person}{Xin Yao}.} \bibinfo{year}{2013}\natexlab{}.
\newblock \showarticletitle{A learning framework for online class imbalance
  learning}. In \bibinfo{booktitle}{\emph{2013 IEEE Symposium on Computational
  Intelligence and Ensemble Learning (CIEL)}}. IEEE, \bibinfo{pages}{36--45}.
\newblock


\bibitem[Wang et~al\mbox{.}(2014)]%
        {wang2014resampling}
\bibfield{author}{\bibinfo{person}{Shuo Wang}, \bibinfo{person}{Leandro~L
  Minku}, {and} \bibinfo{person}{Xin Yao}.} \bibinfo{year}{2014}\natexlab{}.
\newblock \showarticletitle{Resampling-based ensemble methods for online class
  imbalance learning}.
\newblock \bibinfo{journal}{\emph{IEEE Transactions on Knowledge and Data
  Engineering}} \bibinfo{volume}{27}, \bibinfo{number}{5}
  (\bibinfo{year}{2014}), \bibinfo{pages}{1356--1368}.
\newblock


\bibitem[Wang et~al\mbox{.}(2023)]%
        {wang2023towards}
\bibfield{author}{\bibinfo{person}{Zichong Wang}, \bibinfo{person}{Yang Zhou},
  {et~al\mbox{.}}} \bibinfo{year}{2023}\natexlab{}.
\newblock \showarticletitle{Towards Fair Machine Learning Software:
  Understanding and Addressing Model Bias Through Counterfactual Thinking}.
\newblock  (\bibinfo{year}{2023}).
\newblock


\bibitem[Weiss(2004)]%
        {weiss2004mining}
\bibfield{author}{\bibinfo{person}{Gary~M Weiss}.}
  \bibinfo{year}{2004}\natexlab{}.
\newblock \showarticletitle{Mining with rarity: a unifying framework}.
\newblock \bibinfo{journal}{\emph{ACM Sigkdd Explorations Newsletter}}
  \bibinfo{volume}{6}, \bibinfo{number}{1} (\bibinfo{year}{2004}),
  \bibinfo{pages}{7--19}.
\newblock


\bibitem[Yan et~al\mbox{.}(2020)]%
        {yan2020fair}
\bibfield{author}{\bibinfo{person}{Shen Yan}, \bibinfo{person}{Hsien-te Kao},
  {and} \bibinfo{person}{Emilio Ferrara}.} \bibinfo{year}{2020}\natexlab{}.
\newblock \showarticletitle{Fair class balancing: Enhancing model fairness
  without observing sensitive attributes}. In
  \bibinfo{booktitle}{\emph{Proceedings of the 29th ACM International
  Conference on Information \& Knowledge Management}}.
  \bibinfo{pages}{1715--1724}.
\newblock


\bibitem[Zafar et~al\mbox{.}(2019)]%
        {zafar2019fairness}
\bibfield{author}{\bibinfo{person}{Muhammad~Bilal Zafar},
  \bibinfo{person}{Isabel Valera}, \bibinfo{person}{Manuel Gomez-Rodriguez},
  {and} \bibinfo{person}{Krishna~P Gummadi}.} \bibinfo{year}{2019}\natexlab{}.
\newblock \showarticletitle{Fairness constraints: A flexible approach for fair
  classification}.
\newblock \bibinfo{journal}{\emph{The Journal of Machine Learning Research}}
  \bibinfo{volume}{20}, \bibinfo{number}{1} (\bibinfo{year}{2019}),
  \bibinfo{pages}{2737--2778}.
\newblock


\bibitem[Zhang et~al\mbox{.}(2020)]%
        {zhang2020deep}
\bibfield{author}{\bibinfo{person}{Mingli Zhang}, \bibinfo{person}{Xin Zhao},
  {et~al\mbox{.}}} \bibinfo{year}{2020}\natexlab{}.
\newblock \showarticletitle{Deep discriminative learning for autism spectrum
  disorder classification}. In \bibinfo{booktitle}{\emph{International
  Conference on Database and Expert Systems Applications}}. Springer,
  \bibinfo{pages}{435--443}.
\newblock


\bibitem[Zhang et~al\mbox{.}(2021)]%
        {zhang2021farf}
\bibfield{author}{\bibinfo{person}{Wenbin Zhang}, \bibinfo{person}{Albert
  Bifet}, \bibinfo{person}{Xiangliang Zhang}, \bibinfo{person}{Jeremy~C Weiss},
  {and} \bibinfo{person}{Wolfgang Nejdl}.} \bibinfo{year}{2021}\natexlab{}.
\newblock \showarticletitle{FARF: A Fair and Adaptive Random Forests
  Classifier}. In \bibinfo{booktitle}{\emph{Pacific-Asia Conference on
  Knowledge Discovery and Data Mining}}. Springer, \bibinfo{pages}{245--256}.
\newblock


\bibitem[Zhang et~al\mbox{.}(2023a)]%
        {zhang2023fairness}
\bibfield{author}{\bibinfo{person}{Wenbin Zhang}, \bibinfo{person}{Tina
  Hernandez-Boussard}, {and} \bibinfo{person}{Jeremy~C Weiss}.}
  \bibinfo{year}{2023}\natexlab{a}.
\newblock \showarticletitle{Censored Fairness through Awareness}. In
  \bibinfo{booktitle}{\emph{Proceedings of the AAAI Conference on Artificial
  Intelligence}}.
\newblock


\bibitem[Zhang et~al\mbox{.}(2023b)]%
        {zhang2023indiv}
\bibfield{author}{\bibinfo{person}{Wenbin Zhang}, \bibinfo{person}{Juyong Kim},
  \bibinfo{person}{Zichong Wang}, \bibinfo{person}{Pradeep Ravikumar}, {and}
  \bibinfo{person}{Jeremy Weiss}.} \bibinfo{year}{2023}\natexlab{b}.
\newblock \showarticletitle{Individual Fairness Guarantee in Learning with
  Censorship}.
\newblock  (\bibinfo{year}{2023}).
\newblock


\bibitem[Zhang and Ntoutsi(2019)]%
        {zhang2019faht}
\bibfield{author}{\bibinfo{person}{Wenbin Zhang} {and} \bibinfo{person}{Eirini
  Ntoutsi}.} \bibinfo{year}{2019}\natexlab{}.
\newblock \showarticletitle{FAHT: an adaptive fairness-aware decision tree
  classifier}.
\newblock \bibinfo{journal}{\emph{arXiv preprint arXiv:1907.07237}}
  (\bibinfo{year}{2019}).
\newblock


\bibitem[Zhang et~al\mbox{.}(2022)]%
        {zhang2022fairness}
\bibfield{author}{\bibinfo{person}{Wenbin Zhang}, \bibinfo{person}{Shimei Pan},
  \bibinfo{person}{Shuigeng Zhou}, \bibinfo{person}{Toby Walsh}, {and}
  \bibinfo{person}{Jeremy~C Weiss}.} \bibinfo{year}{2022}\natexlab{}.
\newblock \showarticletitle{Fairness Amidst Non-IID Graph Data: Current
  Achievements and Future Directions}.
\newblock \bibinfo{journal}{\emph{arXiv preprint arXiv:2202.07170}}
  (\bibinfo{year}{2022}).
\newblock


\bibitem[Zhang and Wang(2017)]%
        {zhang2017hybrid}
\bibfield{author}{\bibinfo{person}{Wenbin Zhang} {and} \bibinfo{person}{Jianwu
  Wang}.} \bibinfo{year}{2017}\natexlab{}.
\newblock \showarticletitle{A hybrid learning framework for imbalanced stream
  classification}. In \bibinfo{booktitle}{\emph{2017 IEEE International
  Congress on Big Data (BigData Congress)}}. IEEE, \bibinfo{pages}{480--487}.
\newblock


\bibitem[Zhang and Wang(2018)]%
        {zhang2018content}
\bibfield{author}{\bibinfo{person}{Wenbin Zhang} {and} \bibinfo{person}{Jianwu
  Wang}.} \bibinfo{year}{2018}\natexlab{}.
\newblock \showarticletitle{Content-bootstrapped collaborative filtering for
  medical article recommendations}. In \bibinfo{booktitle}{\emph{IEEE
  International Conference on Bioinformatics and Biomedicine (BIBM)}}.
\newblock


\bibitem[Zhang and Weiss(2021)]%
        {zhang2021fair}
\bibfield{author}{\bibinfo{person}{Wenbin Zhang} {and} \bibinfo{person}{Jeremy
  Weiss}.} \bibinfo{year}{2021}\natexlab{}.
\newblock \showarticletitle{Fair Decision-making Under Uncertainty}. In
  \bibinfo{booktitle}{\emph{{2021 IEEE International Conference on Data Mining
  (ICDM)}}}. IEEE.
\newblock


\bibitem[Zhang and Weiss(2022a)]%
        {zhang2022kis}
\bibfield{author}{\bibinfo{person}{Wenbin Zhang} {and} \bibinfo{person}{Jeremy
  Weiss}.} \bibinfo{year}{2022}\natexlab{a}.
\newblock \showarticletitle{Fairness with Censorship and Group Constraints}.
\newblock \bibinfo{journal}{\emph{Knowledge and Information Systems}}
  (\bibinfo{year}{2022}).
\newblock


\bibitem[Zhang and Weiss(2022b)]%
        {zhang2022longitudinal}
\bibfield{author}{\bibinfo{person}{Wenbin Zhang} {and}
  \bibinfo{person}{Jeremy~C Weiss}.} \bibinfo{year}{2022}\natexlab{b}.
\newblock \showarticletitle{Longitudinal fairness with censorship}. In
  \bibinfo{booktitle}{\emph{Proceedings of the AAAI Conference on Artificial
  Intelligence}}, Vol.~\bibinfo{volume}{36}. \bibinfo{pages}{12235--12243}.
\newblock


\bibitem[Zliobaite(2015)]%
        {zliobaite2015survey}
\bibfield{author}{\bibinfo{person}{Indre Zliobaite}.}
  \bibinfo{year}{2015}\natexlab{}.
\newblock \showarticletitle{A survey on measuring indirect discrimination in
  machine learning}.
\newblock \bibinfo{journal}{\emph{arXiv preprint arXiv:1511.00148}}
  (\bibinfo{year}{2015}).
\newblock


\end{thebibliography}

%
%
%

\end{document}